\documentclass{article}

\usepackage{amssymb}
\usepackage{siunitx}
\usepackage{colortbl}
\usepackage{multicol}
\usepackage{xcolor}
\usepackage{graphicx}
\usepackage{svg}
\usepackage{times}
\usepackage{array}
\usepackage{booktabs} %
\usepackage{amsmath}
\usepackage{multirow} %
\usepackage{float}
\usepackage{enumitem}

\usepackage{subcaption}
\newcommand{\gcheck}{\textcolor{green!45!black}{\checkmark}}
\newcommand{\rcross}{\textcolor{red!60!black}{\boldmath$\times$}}

\newcommand{\credit}[1]{\textcolor{black}{#1}}
\newcommand{\creditmm}[1]{\textcolor{black}{#1}}

\newcommand{\PAR}[1]{\vskip2pt \noindent{\bf #1}}

\usepackage[final]{corl_2026} %

\usepackage[capitalise]{cleveref}

\crefname{appendix}{App.}{Apps.}
\Crefname{appendix}{App.}{Apps.}

\title{GoalDex: Learning Image-Goal-Conditioned Policies for Dexterous In-Hand Manipulation}

\author{
  Arjun Bhardwaj \\
  ETH Zurich\\
  \texttt{abhardwaj@ethz.ch} \\
  \And
  Mayank Mittal \\
  ETH Zurich \& NVIDIA\\
  \texttt{mittalma@ethz.ch} \\
  \And
  Maximum Wilder-Smith \\
  ETH Zurich\\
  \texttt{mwilder@ethz.ch} \\
  \And
  Vaishakh Patil \\
  ETH Zurich\\
  \texttt{patilv@ethz.ch} \\
  \And
  Marco Hutter \\
  ETH Zurich\\
  \texttt{mahutter@ethz.ch} \\
}

\begin{document}
\maketitle

\vspace{-16pt}
\begin{abstract}
Dexterous in-hand manipulation is commonly formulated around explicit object pose estimation, yet reliable pose tracking is challenging under severe hand-object occlusions and often requires object-specific supervision. This dependence creates a bottleneck between perception and control that can limit robustness and scalability.
We present an end-to-end visuomotor framework for dexterous in-hand reorientation that operates on RGB observations and proprioception without intermediate pose estimation. Our approach jointly learns visual representations and control policies through behavior distillation from privileged-state teachers and auxiliary reconstruction objectives. Furthermore, we introduce image-goal-conditioned reorientation, allowing desired object configurations to be specified directly through images rather than canonical pose representations.
We evaluate our approach across a diverse set of object geometries and instances in simulation, where it consistently outperforms pose-based methods and policies built on pretrained visual encoders. Finally, we demonstrate direct sim-to-real transfer of a single image-goal-conditioned policy across five real-world objects, showing that accurate dexterous reorientation can be achieved without explicit pose estimation. 
For videos and code release, please refer to the project website: \url{https://rffr.leggedrobotics.com/works/goaldex/}
\end{abstract}


\begin{center}
    \includegraphics[width=\textwidth]{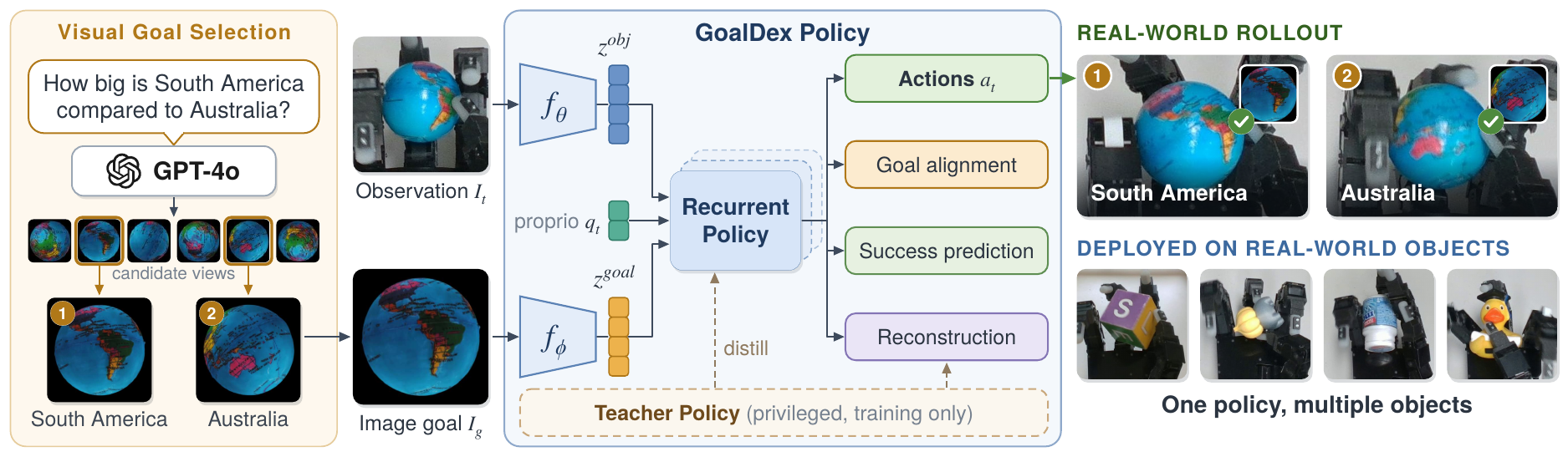}
    \vspace{-16pt}
    \captionof{figure}{
Overview of the proposed image-goal-conditioned framework for dexterous in-hand reorientation.
The policy maps RGB observations, proprioception, and visual goal images directly to joint actions, eliminating the need for explicit pose estimators or canonical object-frame representations.
Trained in simulation, the policy transfers zero-shot to real hardware and is deployed across five diverse objects.
We demonstrate language-guided reorientation using a VLM to select visual goals from natural language instructions.
}
\vspace{-2pt}
    \label{fig:teaser}
\end{center}

\section{Introduction}
\label{sec:introduction}

Manipulating objects into precise functional poses requires a high degree of dexterity. To succeed, a robot must infer object geometry, contact state, and motion despite severe hand-object occlusions. Most existing systems address this challenge through an explicit object pose interface that connects perception and control~\cite{handa2023dextreme,bhardwaj2026viserdex}.
While effective for individual objects, this abstraction introduces important limitations. Object pose discards rich information about local geometry and contact interactions.
It also assumes a predefined canonical coordinate system for every object instance, both in the observation space and in the goal specification.
As manipulation systems scale to diverse or unseen objects, defining such canonical frames becomes increasingly difficult and often ill-posed~\cite{suresh2024neuralfeels}.
More fundamentally, object pose alone cannot capture the transient contact dynamics and local surface features that govern dexterous manipulation. These challenges motivate visuomotor representations that are coordinate-free, temporally grounded, and learned directly from sensory observations.

End-to-end visual control has historically been difficult because pixel observations lack direct physical grounding~\cite{levine2016end}. Recent advances in reinforcement learning~\citep{andrychowicz2020learning}, visual rendering~\citep{mittal2025isaaclab}, and sim-to-real transfer~\cite{bhardwaj2026viserdex} now make it possible to train RGB-based policies in simulation and deploy them on real hardware. A natural approach is to use pretrained visual encoders, which provide strong semantic and geometric priors~\cite{oquab2023dinov2,depth_anything_v2}. However, these models are trained for static visual understanding, not closed-loop physical interaction. Dexterous manipulation requires representations that capture contact, motion, and task-relevant geometry over time. 
These properties may not be readily accessible from generic visual features, motivating representations learned through interaction.

In this work, we eliminate the need for object pose estimation in dexterous in-hand reorientation. Our policy operates on RGB observations and proprioception, with goals specified directly through images. This removes the dependence on object-specific canonical coordinate frames and explicit pose representations.
We develop an end-to-end visuomotor framework that learns coordinate-free, interaction-aware representations from sensory observations. The framework combines privileged teacher–student distillation with a reconstruction objective on hidden system variables, encouraging the latent representation to capture object geometry, contact dynamics, and other task-relevant physical properties.
\credit{Image-goal conditioning also connects the semantic reasoning capabilities of VLMs~\citep{achiam2023gpt} with dexterous control. We demonstrate this connection by using a VLM to select goal images from natural-language instructions.}
Our contributions are as follows:
\begin{itemize}[leftmargin=2em, labelsep=0.4em, itemsep=3pt, parsep=0pt, topsep=0pt]
    \item We present \emph{GoalDex}, an image-goal-conditioned framework for multi-object in-hand reorientation. A single policy maps monocular RGB observations, proprioception, and goal images directly to joint targets, without explicit object-pose inputs or object-frame goal coordinates.

    \item We show that jointly learned visuomotor policies outperform pose-based pipelines and policies using frozen pretrained features. Our ablations quantify the benefits of temporal integration and auxiliary state reconstruction. We further scale training to 25 diverse objects in simulation.

    \item We achieve direct sim-to-real transfer of a single policy trained in simulation on five objects to their real-world counterparts, with 80\% success over 100 hardware trials compared with 27\% for the strongest evaluated pose-based baseline.

\end{itemize}

\section{Related Work}
\label{sec:related_works}

\PAR{Goal-conditioned Dexterous Manipulation.}
Goal-conditioned dexterous manipulation commonly relies on explicit object-state representations to specify goals and provide feedback during execution. Tactile~\citep{rostel2023estimator,pitz2024learning} and RGB-based~\citep{handa2023dextreme,bhardwaj2026viserdex} in-hand reorientation methods use learned state estimators to track object orientation and condition policies on orientation-based goal representations. Recent works further extend this paradigm to 6D goal poses~\citep{kedia2026simtoolrealobjectcentricpolicyzeroshot}, relying on external perception modules~\cite{Wen2023FoundationPoseU6} to estimate the object pose. This dependence on explicit state estimators and object-centric coordinate frames limits scalability to diverse object geometries and unseen instances. Some works have explored visual goal representations, including pixel-based goals, goal latent embeddings~\cite{nair2018rig}, and goal classifiers~\citep{visualforesight}, as well as depth- and point-cloud-based representations for dexterous reorientation~\citep{chen2023visual}. However, direct RGB-based goal conditioning requires learning task-relevant representations directly from high-dimensional observations rather than relying on predefined geometric state representations. In contrast, we formulate dexterous reorientation directly in image space, eliminating the need for object pose estimation and canonical object coordinate frames.

\PAR{Pixel-to-action Visuomotor Learning.}
Pixel-to-action policies have recently shown strong sim-to-real transfer for contact-rich manipulation, reducing the need for explicit state estimation at deployment. DextrAH-G~\citep{lum2024dextrah} trains a depth-based grasping policy using teacher-student distillation. Related works extend this paradigm to more complex embodiments: Lin et al.~\citep{lin2025sim} perform vision-based sim-to-real RL for bimanual dexterous grasping using sparse and dense object representations, while HERMES~\citep{yuan2025hermes} distills state-based policies into depth-image-based policies for mobile bimanual dexterous manipulation. More recently, advances in photorealistic rendering have enabled RGB-based pixel-to-action policies. DextrAH-RGB~\citep{singh2024dextrah} trains end-to-end grasping controllers in simulation that transfer directly to real hardware. Similar RGB distillation paradigms have also been explored for humanoid grasping~\citep{he2025viral} and door traversing~\citep{xue2025opening}. While these works demonstrate the effectiveness of image-based policy distillation, they typically assume fixed task objectives or externally specified goals. Our work extends this paradigm to image-goal-conditioned dexterous manipulation, where both the current state and target configuration are represented in image space.

\section{Method}
\label{sec:method}

We formulate dexterous in-hand reorientation directly in image space. This removes the need for object-specific canonical coordinate frames and allows target configurations to be specified directly through images. We realize this capability through an end-to-end visuomotor framework trained with privileged teacher-student distillation~\citep{takahiroperceptivelocomotion}.

\vspace{-6pt}
\subsection{Problem Formulation}
\label{sec:problem_formulation}
\vspace{-6pt}

We consider the task of in-hand reorientation from an arbitrary initial configuration to a target orientation (in $SO(3)$) specified by a goal image $I_g$.
During training, we assume access to a privileged simulation environment modeled as a goal-conditioned Markov Decision Process (MDP), $\mathcal{M}_{priv} = \langle \mathcal{S}, \mathcal{A}, \mathcal{G}, \mathcal{T}, r, \gamma \rangle$, where \(\mathcal{G} = SO(3)\) denotes the space of possible goals, and $\mathcal{S}$ denotes the privileged state space that encompasses full system dynamics parameters, including exact object geometries, masses, friction coefficients, scales, and spatial poses.
At deployment, the policy operates in a partially-observed setting modeled as a Partially Observable MDP (POMDP), $\mathcal{M}_{sens}$. The deployable policy does not have access to the privileged state \(s_t \in \mathcal{S}\) and must instead map local observations \(o_t \in \mathcal{O}\) to joint targets \(a_t \in \mathcal{A}\). The observation space~$\mathcal{O}$ is multimodal, consisting of a monocular camera stream capturing target-centric RGB frames $I_t$, and a low-level proprioceptive vector $p_t$ encoding joint states and action histories. This formulation is partially observable due to extreme, time-varying finger occlusions, motion blur, and unknown physical dynamics. 

To bridge the sim-to-real gap, we adopt a two-stage teacher-student optimization paradigm. A privileged teacher policy is first optimized within $\mathcal{M}_{priv}$ using full state information.
We then train an end-to-end visuomotor student policy to imitate the teacher while observing only RGB images and proprioception. The resulting policy can be deployed directly on real hardware without access to privileged state information. An overview of the training pipeline is shown in~\cref{fig:pipeline}.

\vspace{-6pt}
\subsection{Teacher Policy Learning}
\label{sec:teacher_learning}
\vspace{-6pt}

The privileged teacher policy $\pi_T(a_t | s_t, g_{pose})$ is trained via Proximal Policy Optimization (PPO)~\cite{Schulman2017ProximalPO} to maximize the expected discounted return. During this phase, the reorientation objective is defined in the geometric domain, where the goal $g_{pose} \in \mathcal{G}$ dictates the target orientation relative to the world coordinate system. The teacher's state vector $s_t$ is partitioned into four distinct categories: $ s_t = \left[ O_{prop}, O_{state}, O_{goal}, O_{class} \right]$. $O_{prop}$ denotes the current robot configuration and action history, $O_{state}$ contains the exact object coordinates and domain-randomized physical parameters (e.g., mass, surface friction), and $O_{goal} \in SO(3)$ defines the instantaneous relative rotation $q_{rel}$ mapping the current orientation to the target frame. To enable robust multi-object capability within a single parameter space, we augment the state vector with a one-hot object identity encoding $O_{class}$. We empirically found this categorical embedding to be necessary for stable teacher policy learning across geometrically diverse, non-convex object instances.

\vspace{-6pt}
\subsection{End-to-End Visuomotor Student Learning}
\label{sec:student_learning}
\vspace{-6pt}

The student policy $\pi_S(a_t \mid h_t)$ must execute dexterous behaviors without access to privileged states, i.e., it does not have access to $O_{state}$, $O_{goal}$, and $O_{class}$. Instead, it must construct an internal interaction-aware representation directly from raw observations. The full architecture for the end-to-end policy distillation consists of the following components.

\PAR{Multimodal Visuomotor Encoding:}
At each execution interval $t$, the student processes the target observation frame $I_t \in \mathbb{R}^{H \times W \times 3}$ using a vision backbone network $f_{\theta}$, providing a compact spatial feature embedding $z_t^{obj} = f_{\theta}(I_t)$. In parallel, a lightweight multi-layer perceptron maps the proprioceptive data $p_t$ to a continuous \mbox{vector~$z_t^{prop}$}. These representations are projected into a unified sensorimotor feature descriptor $\hat{x}_t$ via direct concatenation: $\hat{x}_t = [ z_t^{obj}, z_t^{prop} ]$.

\PAR{Temporal Latent Integration:}
While $x_t$ encodes instantaneous geometric and kinematic features, a single sensor frame cannot resolve transient contact forces, velocity vectors, or unobserved physical dynamics. To maintain an uninterrupted tracking history of the system state, we introduce a recurrent memory module. A multi-layer Recurrent Neural Network~(RNN) processes the multimodal input features sequentially, maintaining a persistent internal belief state $h_t = \text{RNN}(x_t, h_{t-1})$.
The belief state $h_t \in \mathbb{R}^d$ aggregates the temporal context across variable horizons, to implicitly track the object's motion through occlusions, estimate contact dynamics, and accumulate evidence about the object's physical properties. Finally, the updated vector $h_t$ is mapped directly to continuous joint command distributions via an action prediction head.

\PAR{Latent State Reconstruction:}
Optimizing the student policy strictly through the behavior cloning loss $\mathcal{L}_{BC}$ can cause the hidden state $h_t$ to discard physically rich features that do not immediately influence the instantaneous action space. This representational collapse compromises policy robustness under unseen external disturbances. To force the hidden state to build a robust representation of the system dynamics, we introduce an auxiliary reconstruction task. A predictive head $g_{\psi}$ is trained to explicitly reconstruct a critical subset of unobserved variables ${O}_{priv} \subset O_{state}$ directly from the hidden states of the RNN module: $\mathcal{L}_{recon} = \left\| g_{\psi}(h_t) - O_{priv} \right\|_2^2$. By forcing the network to decode physical properties (such as instantaneous velocities and geometric dimensions) from $h_t$, the encoder is incentivized to retain rich physical features alongside the pure control signals.

\vspace{-6pt}
\subsection{Image-Space Goal Conditioning}
\label{sec:image_goal_conditioning}
\label{sec:goal_repr}
\vspace{-6pt}

\begin{figure}[t]
    \centering
    \includegraphics[width=\linewidth]{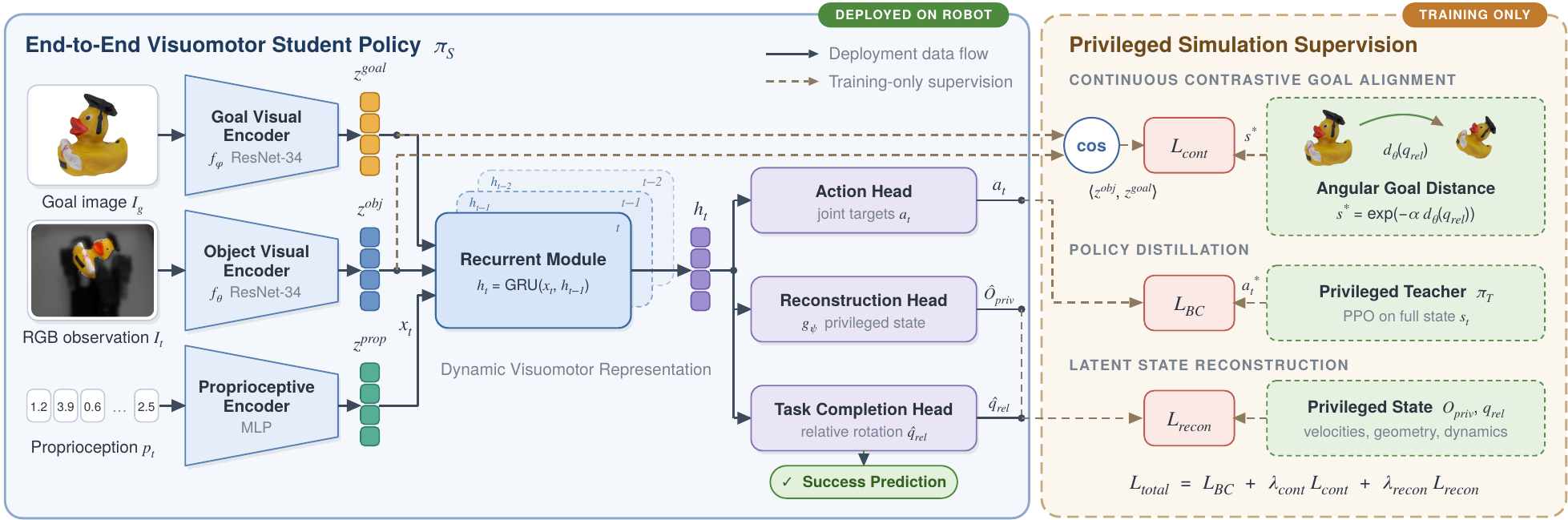}
    \caption{
Training architecture for image-goal-conditioned visuomotor distillation. Current RGB observations and goal images are encoded into visual representations and fused with proprioceptive inputs. A recurrent policy integrates the multimodal features over time to predict joint position targets, while auxiliary reconstruction and continuous contrastive objectives shape the latent representation using privileged simulation supervision.
}
    \label{fig:pipeline}
\vspace{-16pt}
\end{figure}

To avoid reference-frame dependencies during execution, we formulate the orientation tracking problem in image space. The student policy is conditioned on a static goal frame $I_g \in \mathbb{R}^{H \times W \times 3}$ captured from the target configuration. A secondary visual encoder $f_{\phi}$ maps the goal image to a dedicated goal latent embedding: $z^{goal} = f_{\phi}(I_g)$. The complete multimodal input vector feeding the recurrent unit is extended to include this visual target anchor $x_t = [z_t^{obj}, z^{goal}, z_t^{prop}]$.

\PAR{Continuous Contrastive Goal Alignment:}
For an effective goal-conditioned manipulation, the visual embedding space must exhibit geometric continuity, i.e., a small change in the relative rotation of the physical object should map to proportional, predictable distances in the latent space. To achieve this, we introduce a Continuous Contrastive alignment loss that binds the visual representational distances to exact rotational errors. Let $d_{\theta}(q_{rel})$ represent the true angular geodesic distance between the current object pose and the target goal orientation. We define a continuous similarity score $s^* = \exp(-\alpha \cdot d_{\theta}(q_{rel}))$
where $\alpha$ is a scaling hyperparameter. The visual encoders are additionally supervised by minimizing the mean squared error between the continuous similarity score and the cosine similarity of their respective visual outputs: $\mathcal{L}_{cont} = \text{MSE}(s^*, \, \langle z_t^{obj}, z^{goal}\rangle)$, where $\langle \cdot, \cdot \rangle$ is the cosine similarity function. This formulation encourages visual representations to form a coherent metric space where visual distances vary monotonically with physical errors.

\PAR{Success Prediction:}
Since the pipeline operates entirely in the image space without explicitly tracking object pose, defining task success cannot be based on pose error checks. To provide a robust success metric, a task completion head is tasked with directly outputting the camera-frame relative orientation quaternion $\hat{q}_{rel}$. Because this relative calculation is anchored strictly within the camera view matrix rather than an instance-specific canonical coordinate frame, it remains completely category-agnostic. The joint optimization loss function for the entire student network is structured as: $\mathcal{L}_{total} = \mathcal{L}_{BC} + \lambda_{recon}\mathcal{L}_{recon} + \lambda_{cont}\mathcal{L}_{cont}$.

\paragraph{Training:} We train and evaluate our approach on the object set introduced in~\citep{bhardwaj2026viserdex}. The objects have diverse geometric and physical properties ranging from primitive shapes (\emph{Cube}, \emph{Globe}) to complex non-convex objects (\emph{Tablet Bottle}, \emph{3D Printed Toy}, \emph{Rubber Duck}). \credit{Prior work~\citep{bhardwaj2026viserdex} trains a separate policy for each object.} We train a single visuomotor control policy across the entire object set using NVIDIA Isaac Lab~\cite{mittal2025isaaclab}. \creditmm{We sample the object scale factor from $\mathcal{U}(0.5, 1.5)$ independently for each axis. This substantially broadens the range $0.95$--$1.05$ used in~\citep{bhardwaj2026viserdex}. The policy must therefore handle greater variation in object size and visual appearance.}
During training and deployment, the visual encoder receives object-masked RGB observations that isolate the manipulated object from the robot and the background, similar to the prior work. Additional implementation details are in \cref{app:method-details}.

\section{Experiments and Results}
\label{sec:result}

We evaluate our approach through simulated and real-world experiments studying visuomotor representations for in-hand reorientation. We evaluate the policies along two axes: task performance and privileged-state reconstruction. For reconstruction, all students are evaluated on teacher-action rollouts to ensure a consistent state-action distribution across baselines. For task performance, each student controls the environment with their own actions. Simulation results are reported over three random seeds, each with 64  environments and two evaluation episodes per environment.

Reconstruction quality is measured using prediction error over privileged object-state variables. For task performance, we report consecutive successes (CS), time-to-success (TTS), prediction failure rate, and success prediction F1 score. A goal is counted as successful only when the policy predicts completion and the true orientation error is below \(0.4\) radians, following prior work~\citep{handa2023dextreme,andrychowicz2020learning}. Prediction failures correspond to cases where the policy predicts success while the true orientation error remains above the threshold; the current goal is then discarded and a new target is sampled. The F1 score evaluates predicted versus ground-truth success labels at every timestep, while TTS measures the average time required for successful goal completion.

\vspace{-3pt}
\subsection{Sensory Representations for Dexterous Control}
\vspace{-2pt}

\providecommand{\colspacing}{3pt}
\begin{table}[t]
\centering
\renewcommand{\arraystretch}{1.55}
\resizebox{\textwidth}{!}{%
  \setlength{\tabcolsep}{\colspacing} %
  \footnotesize %
  \begin{tabular}{lccccc|cccc|ccccc}
\hline
\multicolumn{6}{c|}{}
& \multicolumn{4}{c|}{\textbf{Task Performance}}
& \multicolumn{5}{c}{\textbf{Privileged-State Reconstruction Error}} \\
\cline{7-15}

\shortstack{\rule{0pt}{2.5ex}~ \\ JP \\ ~\rule[-1.5ex]{0pt}{0pt}} & \shortstack{\rule{0pt}{2.5ex}~ \\ P \\ ~\rule[-1.5ex]{0pt}{0pt}} & \shortstack{\rule{0pt}{2.5ex}~ \\ C \\ ~\rule[-1.5ex]{0pt}{0pt}} & \shortstack{\rule{0pt}{2.5ex}~ \\ I \\ ~\rule[-1.5ex]{0pt}{0pt}} & \shortstack{\rule{0pt}{2.5ex}~ \\ GP \\ ~\rule[-1.5ex]{0pt}{0pt}} & \shortstack{\rule{0pt}{2.5ex}~ \\ GI \\ ~\rule[-1.5ex]{0pt}{0pt}} & \shortstack{\rule{0pt}{2.5ex}\textbf{Consecutive} \\ \textbf{Successes} \\ $\uparrow$\rule[-1.5ex]{0pt}{0pt}} & \shortstack{\rule{0pt}{2.5ex}\textbf{Time to} \\  \textbf{Success} \\ \textbf{(s)}~$\downarrow$\rule[-1.5ex]{0pt}{0pt}} & \shortstack{\rule{0pt}{2.5ex}\textbf{Success} \\ \textbf{Prediction} \\ \textbf{F1 Score}~$\uparrow$\rule[-1.5ex]{0pt}{0pt}} & \shortstack{\rule{0pt}{2.5ex}\textbf{Prediction} \\ \textbf{Failure} \\ \textbf{Rate (\%)}~$\downarrow$\rule[-1.5ex]{0pt}{0pt}} & \shortstack{\rule{0pt}{2.5ex}Object \\ Position \\ (mm)~$\downarrow$\rule[-1.5ex]{0pt}{0pt}} & \shortstack{\rule{0pt}{2.5ex}Object \\ Orientation \\ (rad)~$\downarrow$\rule[-1.5ex]{0pt}{0pt}} & \shortstack{\rule{0pt}{2.5ex}Relative \\ Orientation \\ (rad)~$\downarrow$\rule[-1.5ex]{0pt}{0pt}} & \shortstack{\rule{0pt}{2.5ex}Object \\ Scale \\ (\%)~$\downarrow$\rule[-1.5ex]{0pt}{0pt}} & \shortstack{\rule{0pt}{2.5ex}Object \\ Mass \\ (\%)~$\downarrow$\rule[-1.5ex]{0pt}{0pt}} \\
\hline
\multicolumn{15}{l}{\emph{Proprioception Only Baseline}} \\
\gcheck & \rcross & \rcross & \rcross & \gcheck & \rcross & \cellcolor[HTML]{EECDD3} $1.05_{\pm 0.00}$ & \cellcolor[HTML]{EDCCD3} $8.31_{\pm 0.00}$ & \cellcolor[HTML]{EECDD3} $0.016_{\pm 0.000}$ & \cellcolor[HTML]{EDCCD3} $99.18_{\pm 0.00}$ & \cellcolor[HTML]{EDCCD3} $18.83_{\pm 0.12}$ & \cellcolor[HTML]{EDCCD3} $2.220_{\pm 0.017}$ & \cellcolor[HTML]{EDCCD3} $1.177_{\pm 0.007}$ & \cellcolor[HTML]{EDCCD3} $10.6_{\pm 0.1}$ & \cellcolor[HTML]{FEF4E4} $33.2_{\pm 0.7}$ \\
\hline
\multicolumn{15}{l}{\emph{Pose-Based Baselines \credit{(extends \cite{bhardwaj2026viserdex} to multiple objects)}}}\ \\
\gcheck & \gcheck & \rcross & \rcross & \gcheck & \rcross & \cellcolor[HTML]{F7D6D4} $8.64_{\pm 10.17}$ & \cellcolor[HTML]{F5D4D3} $7.96_{\pm 0.53}$ & \cellcolor[HTML]{FEF9E8} $0.373_{\pm 0.127}$ & \cellcolor[HTML]{FBFDED} $44.77_{\pm 2.77}$ & \cellcolor[HTML]{CFE6DA} $5.40_{\pm 0.36}$ & \cellcolor[HTML]{FEFEF1} $1.169_{\pm 0.041}$ & \cellcolor[HTML]{F3FAE5} $0.517_{\pm 0.054}$ & \cellcolor[HTML]{FEECDE} $8.6_{\pm 0.4}$ & \cellcolor[HTML]{DBEFDE} $24.8_{\pm 1.3}$ \\
\gcheck & \gcheck & \gcheck & \rcross & \gcheck & \rcross & \cellcolor[HTML]{F9DBD6} $11.23_{\pm 9.65}$ & \cellcolor[HTML]{F9DBD6} $7.73_{\pm 0.74}$ & \cellcolor[HTML]{FEFDEF} $0.439_{\pm 0.111}$ & \cellcolor[HTML]{FAFDEB} $43.76_{\pm 3.32}$ & \cellcolor[HTML]{CFE6DA} $5.43_{\pm 0.06}$ & \cellcolor[HTML]{FEEBDD} $1.654_{\pm 0.255}$ & \cellcolor[HTML]{FBFDED} $0.614_{\pm 0.036}$ & \cellcolor[HTML]{FDE6DB} $8.9_{\pm 0.4}$ & \cellcolor[HTML]{D0E8DB} $\underline{23.3_{\pm 1.1}}$ \\
\rcross & \gcheck & \rcross & \rcross & \gcheck & \rcross & \cellcolor[HTML]{EECDD3} $1.81_{\pm 0.65}$ & \cellcolor[HTML]{F5D4D3} $7.96_{\pm 0.44}$ & \cellcolor[HTML]{FEFAEA} $0.394_{\pm 0.092}$ & \cellcolor[HTML]{FEFEF1} $48.63_{\pm 3.48}$ & \cellcolor[HTML]{D2EADC} $6.03_{\pm 0.23}$ & \cellcolor[HTML]{F8FCE9} $1.004_{\pm 0.132}$ & \cellcolor[HTML]{EAF6E0} $0.428_{\pm 0.045}$ & \cellcolor[HTML]{FEF2E2} $8.2_{\pm 0.1}$ & \cellcolor[HTML]{CCE0D7} $\mathbf{21.8_{\pm 1.5}}$ \\
\rcross & \gcheck & \gcheck & \rcross & \gcheck & \rcross & \cellcolor[HTML]{F2D0D3} $4.08_{\pm 1.46}$ & \cellcolor[HTML]{F7D7D4} $7.86_{\pm 0.40}$ & \cellcolor[HTML]{FEFAEA} $0.396_{\pm 0.065}$ & \cellcolor[HTML]{FDFEEF} $47.26_{\pm 3.38}$ & \cellcolor[HTML]{CFE6DA} $5.40_{\pm 0.10}$ & \cellcolor[HTML]{F7FBE7} $0.974_{\pm 0.098}$ & \cellcolor[HTML]{EFF8E2} $0.475_{\pm 0.021}$ & \cellcolor[HTML]{FBDFD8} $9.3_{\pm 0.2}$ & \cellcolor[HTML]{D2EBDC} $23.8_{\pm 0.1}$ \\
\hline
\multicolumn{15}{l}{\emph{Image-Based Policies (Ours)}} \\
\rcross & \rcross & \rcross & \gcheck & \gcheck & \rcross & \cellcolor[HTML]{EDCCD3} $0.01_{\pm 0.01}$ & \cellcolor[HTML]{FEF6E6} $6.80_{\pm 0.22}$ & \cellcolor[HTML]{EDCCD3} $0.000_{\pm 0.000}$ & \cellcolor[HTML]{EDCCD3} $100.00_{\pm 0.00}$ & \cellcolor[HTML]{D0E9DB} $5.89_{\pm 0.16}$ & \cellcolor[HTML]{E6F4E0} $0.676_{\pm 0.013}$ & \cellcolor[HTML]{FDFEF0} $0.637_{\pm 0.005}$ & \cellcolor[HTML]{EFCED3} $10.4_{\pm 0.2}$ & \cellcolor[HTML]{EDCCD3} $39.7_{\pm 1.5}$ \\
\gcheck & \rcross & \rcross & \gcheck & \gcheck & \rcross & \cellcolor[HTML]{D1EADC} $\underline{69.20_{\pm 5.47}}$ & \cellcolor[HTML]{E1F2DF} $\underline{5.17_{\pm 0.40}}$ & \cellcolor[HTML]{CCE0D7} $\mathbf{0.923_{\pm 0.005}}$ & \cellcolor[HTML]{D0E8DB} $\mathbf{8.33_{\pm 0.07}}$ & \cellcolor[HTML]{CCE1D7} $\underline{4.70_{\pm 0.00}}$ & \cellcolor[HTML]{CCE1D7} $\underline{0.165_{\pm 0.004}}$ & \cellcolor[HTML]{CCE1D7} $\underline{0.152_{\pm 0.003}}$ & \cellcolor[HTML]{FEF8E7} $\underline{7.8_{\pm 0.1}}$ & \cellcolor[HTML]{F2F9E4} $28.1_{\pm 0.4}$ \\
\hline
\multicolumn{15}{l}{\emph{Image-Goal Conditioned (Ours)}} \\
\gcheck & \rcross & \rcross & \gcheck & \rcross & \gcheck & \cellcolor[HTML]{CCE0D7} $\mathbf{77.30_{\pm 3.25}}$ & \cellcolor[HTML]{CCE0D7} $\mathbf{4.35_{\pm 0.24}}$ & \cellcolor[HTML]{CCE0D7} $\underline{0.921_{\pm 0.000}}$ & \cellcolor[HTML]{D0E9DB} \underline{$8.80_{\pm 0.15}$} & \cellcolor[HTML]{CCE0D7} $\mathbf{4.50_{\pm 0.00}}$ & \cellcolor[HTML]{CCE0D7} $\mathbf{0.151_{\pm 0.004}}$ & \cellcolor[HTML]{CCE0D7} $\mathbf{0.136_{\pm 0.003}}$ & \cellcolor[HTML]{CCE0D7} $\mathbf{3.5_{\pm 0.1}}$ & \cellcolor[HTML]{DAEEDE} $24.7_{\pm 2.6}$ \\
\hline
\vspace{-0.35cm}
\end{tabular}%
}
\caption{
Effect of sensory inputs and goal representations.
JP: joint proprioception; P: predicted object pose; C: object class encoding;
I: current RGB image; GP: pose-space goal; GI: image-space goal.
}
\label{tab:metrics_modality_ablation}
\vspace{-14pt}
\end{table}

We evaluate different combinations of raw sensory observations and pose-based inputs for in-hand reorientation. All student policies share the same architecture, differing only in the dimensionality of their input layers. For pose-based baselines, we pre-train a pose estimator on rollouts generated by the teacher policy using the same object set. To ensure a fair comparison, the pose estimator uses the same visual backbone (ResNet-34~\cite{He2015DeepRL}) as the visual encoder employed by the image-based student policies, resulting in comparable model capacity across methods. \credit{More details are in \cref{app:baselines}.}

\cref{tab:metrics_modality_ablation} evaluates the impact of different sensing and goal modalities on in-hand reorientation. Pose-based baselines achieve limited task performance, even when object identity is provided, suggesting that pose information alone is insufficient for robust multi-object control. \credit{These baselines also show higher reconstruction errors. These errors should not be interpreted as pose-estimator failures. The pose errors reported in \cref{tab:metrics_modality_ablation} arise from the policy's reconstruction head rather than the pose estimator itself. The pose estimator, evaluated standalone, achieves a mean position error of 6.8~mm and a mean orientation error of 0.25~rad (\cref{sec:pose_estimator_evaluation}). Thus, these errors reflect inaccuracies in the policy's reconstructed belief state, rather than inaccuracies in the pose estimator.}

Using image observations without proprioception also fails, showing that vision alone cannot provide the contact feedback required for the task. Combining RGB observations with proprioception substantially improves both reconstruction and task performance, demonstrating the complementary roles of visual geometry and proprioceptive contact feedback. Our image-goal-conditioned policy achieves the strongest overall performance, achieving the highest consecutive successes.
\credit{Importantly, all modalities evaluated in~\cref{tab:metrics_modality_ablation} are available on real robots during deployment. We additionally compare the policy against privileged and oracle-pose policies in \cref{app:oracle-baselines}.}

\vspace{-3pt}
\subsection{Architectures for Dynamic Visuomotor Representations}
\vspace{-2pt}

\providecommand{\colspacing}{3pt}
\begin{table}[t]
\centering
\renewcommand{\arraystretch}{1.5}
\resizebox{\textwidth}{!}{%
  \setlength{\tabcolsep}{\colspacing} %
  \footnotesize %
  \begin{tabular}{l|cccc|ccccc}
\hline
\multicolumn{1}{c|}{}
& \multicolumn{4}{c|}{\textbf{Task Performance}}
& \multicolumn{5}{c}{\textbf{Privileged-State Reconstruction Error}} \\
\cline{2-10}
\shortstack{\rule{0pt}{2.5ex}~ \\ \textbf{Architecture} \\ ~\rule[-1.5ex]{0pt}{0pt}} & \shortstack{\rule{0pt}{2.5ex}\textbf{Consecutive} \\ \textbf{Successes} \\ $\uparrow$\rule[-1.5ex]{0pt}{0pt}} & \shortstack{\rule{0pt}{2.5ex}\textbf{Time to} \\  \textbf{Success} \\ \textbf{(s)}~$\downarrow$\rule[-1.5ex]{0pt}{0pt}} & \shortstack{\rule{0pt}{2.5ex}\textbf{Success} \\ \textbf{Prediction} \\ \textbf{F1 Score}~$\uparrow$\rule[-1.5ex]{0pt}{0pt}} & \shortstack{\rule{0pt}{2.5ex}\textbf{Prediction} \\ \textbf{Failure} \\ \textbf{Rate (\%)}~$\downarrow$\rule[-1.5ex]{0pt}{0pt}} & \shortstack{\rule{0pt}{2.5ex}Object \\ Position \\ (mm)~$\downarrow$\rule[-1.5ex]{0pt}{0pt}} & \shortstack{\rule{0pt}{2.5ex}Object \\ Orientation \\ (rad)~$\downarrow$\rule[-1.5ex]{0pt}{0pt}} & \shortstack{\rule{0pt}{2.5ex}Relative \\ Orientation \\ (rad)~$\downarrow$\rule[-1.5ex]{0pt}{0pt}} & \shortstack{\rule{0pt}{2.5ex}Object \\ Scale \\ (\%)~$\downarrow$\rule[-1.5ex]{0pt}{0pt}} & \shortstack{\rule{0pt}{2.5ex}Object \\ Mass \\ (\%)~$\downarrow$\rule[-1.5ex]{0pt}{0pt}} \\
\hline
Frozen ImageNet            & \cellcolor[HTML]{EDCCD3} $0.02_{\pm 0.00}$ & \cellcolor[HTML]{FADCD6} $8.18_{\pm 0.49}$ & \cellcolor[HTML]{F0CFD3} $0.029_{\pm 0.003}$ & \cellcolor[HTML]{EECDD3} $98.65_{\pm 0.18}$ & \cellcolor[HTML]{F5FBE6} $11.70_{\pm 0.10}$ & \cellcolor[HTML]{FEFDEF} $1.238_{\pm 0.022}$ & \cellcolor[HTML]{FEF6E5} $1.179_{\pm 0.012}$ & \cellcolor[HTML]{F4D2D3} $9.8_{\pm 0.1}$ & \cellcolor[HTML]{FDFEEF} $30.6_{\pm 3.3}$ \\
Frozen DinoV2              & \cellcolor[HTML]{EDCCD3} $0.00_{\pm 0.00}$ & - & \cellcolor[HTML]{EDCCD3} $0.000_{\pm 0.000}$ & \cellcolor[HTML]{EDCCD3} $100.00_{\pm 0.00}$ & \cellcolor[HTML]{EDCCD3} $23.20_{\pm 0.17}$ & \cellcolor[HTML]{EDCCD3} $2.229_{\pm 0.015}$ & \cellcolor[HTML]{EDCCD3} $1.805_{\pm 0.009}$ & \cellcolor[HTML]{EDCCD3} $10.3_{\pm 0.2}$ & \cellcolor[HTML]{EDCCD3} $39.0_{\pm 3.0}$ \\
Frozen DepthAnythingV2     & \cellcolor[HTML]{EDCCD3} $0.12_{\pm 0.00}$ & \cellcolor[HTML]{F6D4D3} $8.46_{\pm 0.00}$ & \cellcolor[HTML]{F0CFD3} $0.036_{\pm 0.000}$ & \cellcolor[HTML]{F0CFD3} $96.72_{\pm 0.00}$ & \cellcolor[HTML]{F9FCEB} $12.60_{\pm 0.10}$ & \cellcolor[HTML]{FEFBEC} $1.309_{\pm 0.022}$ & \cellcolor[HTML]{FEECDE} $1.335_{\pm 0.017}$ & \cellcolor[HTML]{F8D9D5} $9.4_{\pm 0.6}$ & \cellcolor[HTML]{FEFAEB} $32.0_{\pm 1.5}$ \\
Pose-only Reconstruction   & \cellcolor[HTML]{D8EDDD} $\underline{66.03_{\pm 9.34}}$ & \cellcolor[HTML]{E8F5E0} $5.55_{\pm 0.21}$ & \cellcolor[HTML]{CCE1D7} $0.912_{\pm 0.002}$ & \cellcolor[HTML]{CCE1D7} $\underline{9.70_{\pm 0.70}}$ & \cellcolor[HTML]{CCE0D7} $4.57_{\pm 0.12}$ & \cellcolor[HTML]{CDE2D7} $0.195_{\pm 0.007}$ & \cellcolor[HTML]{CCE1D7} $0.149_{\pm 0.003}$ & - & - \\
No Memory                  & \cellcolor[HTML]{FEFEF1} $38.64_{\pm 1.89}$ & \cellcolor[HTML]{E7F4E0} $\underline{5.51_{\pm 0.28}}$ & \cellcolor[HTML]{CEE5D9} $0.872_{\pm 0.006}$ & \cellcolor[HTML]{CFE6DA} $14.50_{\pm 0.22}$ & \cellcolor[HTML]{CEE4D8} $5.27_{\pm 0.06}$ & \cellcolor[HTML]{CDE3D8} $0.224_{\pm 0.004}$ & \cellcolor[HTML]{CDE2D7} $0.171_{\pm 0.005}$ & \cellcolor[HTML]{FEEFE0} $8.2_{\pm 0.1}$ & \cellcolor[HTML]{D4EBDC} $24.9_{\pm 1.3}$ \\
No Visual Memory           & \cellcolor[HTML]{F5FAE6} $47.83_{\pm 9.07}$ & \cellcolor[HTML]{F2F9E4} $5.93_{\pm 0.35}$ & \cellcolor[HTML]{CCE1D7} $\underline{0.913_{\pm 0.003}}$ & \cellcolor[HTML]{CDE3D8} $11.80_{\pm 1.31}$ & \cellcolor[HTML]{CCE0D7} $\mathbf{4.50_{\pm 0.00}}$ & \cellcolor[HTML]{CCE1D7} $\underline{0.166_{\pm 0.005}}$ & \cellcolor[HTML]{CCE0D7} $\underline{0.142_{\pm 0.002}}$ & \cellcolor[HTML]{FEFCED} $\underline{7.2_{\pm 0.2}}$ & \cellcolor[HTML]{CCE0D7} $\mathbf{22.9_{\pm 1.2}}$ \\
Ours                       & \cellcolor[HTML]{CCE0D7} $\mathbf{77.30_{\pm 3.25}}$ & \cellcolor[HTML]{CCE0D7} $\mathbf{4.35_{\pm 0.24}}$ & \cellcolor[HTML]{CCE0D7} $\mathbf{0.921_{\pm 0.000}}$ & \cellcolor[HTML]{CCE0D7} $\mathbf{8.80_{\pm 0.15}}$ & \cellcolor[HTML]{CCE0D7} $\mathbf{4.50_{\pm 0.00}}$ & \cellcolor[HTML]{CCE0D7} $\mathbf{0.151_{\pm 0.004}}$ & \cellcolor[HTML]{CCE0D7} $\mathbf{0.136_{\pm 0.003}}$ & \cellcolor[HTML]{CCE0D7} $\mathbf{3.5_{\pm 0.1}}$ & \cellcolor[HTML]{D2EBDC} $\underline{24.7_{\pm 2.6}}$ \\
\hline
\end{tabular}%
}
\vspace{0.1cm}
\caption{
Comparison of visual encoding, reconstruction objectives, and temporal memory variants.
}
\label{tab:metrics_representations_ablation}
\vspace{-22pt}
\end{table}

We evaluate whether pretrained visual representations, including \emph{ImageNet-pretrained ResNet} features~\citep{5206848}, \emph{DINOv2}~\citep{oquab2023dinov2}, and \emph{Depth Anything V2}~\citep{depth_anything_v2}, capture sufficient information for dynamic systems. In addition, we perform architectural ablations to study the role of reconstruction learning and temporal memory in learning interaction-aware visuomotor representations. Specifically, we have the following ablations.
\emph{Pose-Only Reconstruction:} The reconstruction objective is restricted to object pose information only, similar to prior RGB-based policy distillation approaches~\citep{singh2024dextrah}.
\emph{No Memory:} We replace the recurrent memory module with feedforward multilayer perceptrons (MLPs) of equivalent size, removing temporal state aggregation.
\emph{No Visual Memory:} The recurrent memory module processes only proprioceptive features, while the visual embeddings \([z_t^{obj}, z^{goal}]\) are provided directly to the policy without temporal aggregation.

We report the results in ~\cref{tab:metrics_representations_ablation}. Frozen pretrained encoders, including ImageNet, DINOv2, and Depth Anything V2, fail on both privileged-state reconstruction and task execution, indicating that generic visual features are poorly aligned with the interaction-relevant information required for the task. Pose-only reconstruction yields non-trivial performance but remains below our full model, suggesting that additional privileged quantities beyond pose provide important supervision for control. We also trained a variant without the auxiliary reconstruction head, while keeping the rest of the architecture unchanged, resulting in complete policy failure.
Similarly, removing recurrent memory degrades both reconstruction and task performance, while removing visual memory preserves strong reconstruction accuracy but substantially reduces consecutive successes. This indicates that temporally integrated visual features are important not only for state prediction, but also for closed-loop control under occlusions and object motion. Overall, our full model achieves the best task performance supporting the need for dynamic, reconstruction-supervised visuomotor representations.

\vspace{-3pt}
\subsection{Image-Goal Conditioning and Representation Alignment}
\vspace{-2pt}

\providecommand{\colspacing}{3pt}
\begin{table}[t]
\centering
\renewcommand{\arraystretch}{1.5}
\resizebox{\textwidth}{!}{%
  \setlength{\tabcolsep}{\colspacing} %
  \footnotesize %
  \begin{tabular}{l|cccc|ccccc}
\hline
& \multicolumn{4}{c|}{\textbf{Task Performance}}
& \multicolumn{5}{c}{\textbf{Privileged-State Reconstruction Error}} \\
\cline{2-10}
\shortstack{\rule{0pt}{2.5ex}~ \textbf{Alignment} \\ \textbf{Strategy} \\ ~\rule[-1.5ex]{0pt}{0pt}} & \shortstack{\rule{0pt}{2.5ex}\textbf{Consecutive} \\ \textbf{Successes} \\ $\uparrow$\rule[-1.5ex]{0pt}{0pt}} & \shortstack{\rule{0pt}{2.5ex}\textbf{Time to} \\  \textbf{Success} \\ \textbf{(s)}~$\downarrow$\rule[-1.5ex]{0pt}{0pt}} & \shortstack{\rule{0pt}{2.5ex}\textbf{Success} \\ \textbf{Prediction} \\ \textbf{F1 Score}~$\uparrow$\rule[-1.5ex]{0pt}{0pt}} & \shortstack{\rule{0pt}{2.5ex}\textbf{Prediction} \\ \textbf{Failure} \\ \textbf{Rate (\%)}~$\downarrow$\rule[-1.5ex]{0pt}{0pt}} & \shortstack{\rule{0pt}{2.5ex}Object \\ Position \\ (mm)~$\downarrow$\rule[-1.5ex]{0pt}{0pt}} & \shortstack{\rule{0pt}{2.5ex}Object \\ Orientation \\ (rad)~$\downarrow$\rule[-1.5ex]{0pt}{0pt}} & \shortstack{\rule{0pt}{2.5ex}Relative \\ Orientation \\ (rad)~$\downarrow$\rule[-1.5ex]{0pt}{0pt}} & \shortstack{\rule{0pt}{2.5ex}Object \\ Scale \\ (\%)~$\downarrow$\rule[-1.5ex]{0pt}{0pt}} & \shortstack{\rule{0pt}{2.5ex}Object \\ Mass \\ (\%)~$\downarrow$\rule[-1.5ex]{0pt}{0pt}} \\
\hline
Separate                   & \cellcolor[HTML]{EDCCD3} $56.99_{\pm 1.89}$ & \cellcolor[HTML]{D3EBDC} $\underline{4.50_{\pm 1.12}}$ & \cellcolor[HTML]{EBF6E1} $0.913_{\pm 0.004}$ & \cellcolor[HTML]{EDCCD3} $10.50_{\pm 0.61}$ & \cellcolor[HTML]{F9DAD6} $4.57_{\pm 0.06}$ & \cellcolor[HTML]{FBFDED} $0.171_{\pm 0.004}$ & \cellcolor[HTML]{EAF6E0} $0.154_{\pm 0.004}$ & \cellcolor[HTML]{FEEEDF} $6.0_{\pm 0.2}$ & \cellcolor[HTML]{F9DAD5} $25.1_{\pm 1.6}$ \\
EMA Updates                & \cellcolor[HTML]{FBFDED} $\underline{67.95_{\pm 6.80}}$ & \cellcolor[HTML]{EDCCD3} $5.65_{\pm 0.40}$ & \cellcolor[HTML]{EDCCD3} $0.857_{\pm 0.004}$ & \cellcolor[HTML]{F5D3D3} $10.30_{\pm 0.58}$ & \cellcolor[HTML]{EDCCD3} $4.67_{\pm 0.12}$ & \cellcolor[HTML]{EDCCD3} $0.196_{\pm 0.001}$ & \cellcolor[HTML]{EDCCD3} $0.208_{\pm 0.003}$ & \cellcolor[HTML]{EDCCD3} $7.4_{\pm 0.1}$ & \cellcolor[HTML]{FEEBDD} $\underline{24.7_{\pm 1.8}}$ \\
Shared                     & \cellcolor[HTML]{FEFEF1} $67.18_{\pm 4.43}$ & \cellcolor[HTML]{E7F5E0} $4.69_{\pm 0.42}$ & \cellcolor[HTML]{CCE0D7} $\mathbf{0.936_{\pm 0.002}}$ & \cellcolor[HTML]{FEFEF1} $9.32_{\pm 0.12}$ & \cellcolor[HTML]{F0F8E3} $\underline{4.20_{\pm 0.00}}$ & \cellcolor[HTML]{CCE0D7} $\mathbf{0.150_{\pm 0.003}}$ & \cellcolor[HTML]{CCE0D7} $\mathbf{0.133_{\pm 0.002}}$ & \cellcolor[HTML]{CCE0D7} $\mathbf{2.7_{\pm 0.2}}$ & \cellcolor[HTML]{EDCCD3} $25.5_{\pm 2.0}$ \\
InfoNCE                    & \cellcolor[HTML]{FEEEDF} $63.01_{\pm 15.51}$ & \cellcolor[HTML]{E4F3E0} $4.65_{\pm 0.37}$ & \cellcolor[HTML]{CDE3D8} $\underline{0.934_{\pm 0.001}}$ & \cellcolor[HTML]{CCE0D7} $\mathbf{8.11_{\pm 0.54}}$ & \cellcolor[HTML]{CCE0D7} $\mathbf{3.97_{\pm 0.06}}$ & \cellcolor[HTML]{E7F5E0} $0.162_{\pm 0.003}$ & \cellcolor[HTML]{CCE0D7} $\mathbf{0.133_{\pm 0.002}}$ & \cellcolor[HTML]{FEFBEC} $5.3_{\pm 0.1}$ & \cellcolor[HTML]{CCE0D7} $\mathbf{22.6_{\pm 0.6}}$ \\
Ours                       & \cellcolor[HTML]{CCE0D7} $\mathbf{77.30_{\pm 3.25}}$ & \cellcolor[HTML]{CCE0D7} $\mathbf{4.35_{\pm 0.24}}$ & \cellcolor[HTML]{DEF0DF} $0.921_{\pm 0.000}$ & \cellcolor[HTML]{EBF6E1} $\underline{8.80_{\pm 0.15}}$ & \cellcolor[HTML]{FDE7DB} $4.50_{\pm 0.00}$ & \cellcolor[HTML]{CDE2D7} $\underline{0.151_{\pm 0.004}}$ & \cellcolor[HTML]{CEE4D8} $0.136_{\pm 0.003}$ & \cellcolor[HTML]{DBEFDE} $\underline{3.5_{\pm 0.1}}$ & \cellcolor[HTML]{FEEBDD} $24.7_{\pm 2.6}$ \\
\hline
\end{tabular}%
}
\vspace{0.1cm}
\caption{
Effect of object--goal alignment on task performance and state reconstruction.
}
\label{tab:metrics_alignment_ablation}
\vspace{-15pt}
\end{table}

\providecommand{\colspacing}{3pt}
\begin{table}[t]
\centering
\renewcommand{\arraystretch}{1.5}
\resizebox{\textwidth}{!}{%
  \setlength{\tabcolsep}{\colspacing} %
  \footnotesize %
  \begin{tabular}{l|cccc|ccccc}
\hline
& \multicolumn{4}{c|}{\textbf{Task Performance}}
& \multicolumn{5}{c}{\textbf{Privileged-State Reconstruction Error}} \\
\cline{2-10}
\textbf{\shortstack{\rule{0pt}{2.5ex}~ \\ Method \\ ~\rule[-1.5ex]{0pt}{0pt}}} &
\textbf{\shortstack{\rule{0pt}{2.5ex}Consecutive \\ Successes \\ $\uparrow$\rule[-1.5ex]{0pt}{0pt}}} &
\textbf{\shortstack{\rule{0pt}{2.5ex}Time to \\ Success \\ (s)~$\downarrow$\rule[-1.5ex]{0pt}{0pt}}} &
\textbf{\shortstack{\rule{0pt}{2.5ex}Success \\ Prediction \\ F1 Score~$\uparrow$\rule[-1.5ex]{0pt}{0pt}}} &
\textbf{\shortstack{\rule{0pt}{2.5ex}Prediction \\ Failure \\ Rate (\%)~$\downarrow$\rule[-1.5ex]{0pt}{0pt}}} &
{\shortstack{\rule{0pt}{2.5ex}Object \\ Position \\ (mm)~$\downarrow$\rule[-1.5ex]{0pt}{0pt}}} &
{\shortstack{\rule{0pt}{2.5ex}Object \\ Orientation \\ (rad)~$\downarrow$\rule[-1.5ex]{0pt}{0pt}}} &
{\shortstack{\rule{0pt}{2.5ex}Relative \\ Orientation \\ (rad)~$\downarrow$\rule[-1.5ex]{0pt}{0pt}}} &
{\shortstack{\rule{0pt}{2.5ex}Object \\ Scale \\ (\%)~$\downarrow$\rule[-1.5ex]{0pt}{0pt}}} &
{\shortstack{\rule{0pt}{2.5ex}Object \\ Mass \\ (\%)~$\downarrow$\rule[-1.5ex]{0pt}{0pt}}} \\
\hline
Privileged Teacher & $\mathbf{110.84_{\pm 3.98}}$ & $\mathbf{2.42_{\pm 0.37}}$ & - & - & - & - & - & - & - \\
\hline
Image-Goal Conditioned & $10.11_{\pm 2.06}$ & $6.74_{\pm 0.05}$ & $0.636_{\pm 0.003}$ & $29.31_{\pm 2.17}$ & $7.13_{\pm 0.06}$ & $0.399_{\pm 0.002}$ & $0.393_{\pm 0.003}$ & $9.8_{\pm 0.4}$ & $31.8_{\pm 2.2}$ \\
Image-Goal Conditioned 2x & $41.42_{\pm 0.00}$ & $5.30_{\pm 0.00}$ & $\mathbf{0.882_{\pm 0.000}}$ & $\mathbf{9.61_{\pm 0.00}}$ & $\mathbf{4.60_{\pm 0.10}}$ & $\mathbf{0.154_{\pm 0.004}}$ & $\mathbf{0.190_{\pm 0.008}}$ & $\mathbf{5.1_{\pm 0.2}}$ & $\mathbf{29.0_{\pm 2.8}}$ \\
\hline
\end{tabular}%
}
\vspace{3pt}
\caption{Scaling to 25 objects in simulation. $2\times$ denotes twice as many parallel training environments.}
\label{tab:metrics_scale}
\vspace{-24pt}
\end{table}

We evaluate the impact of different representation alignment strategies between the object representation \(z_t^{obj}\) and the goal representation \(z^{goal}\) on policy performance and success prediction accuracy.  We compare our continuous contrastive objective, as described in~\cref{sec:goal_repr}, against several alternative alignment strategies. First, we evaluate a baseline without any explicit contrastive alignment objective. We additionally investigate an InfoNCE~\cite{oord2018representation} alignment objective, where positive pairs correspond to object and goal configurations whose relative orientation lies within the success threshold. Finally, we study the impact of encoder sharing strategies by evaluating a fully shared variant that uses the same visual backbone for both object and goal image embeddings, as well as an exponential moving average (EMA) formulation in which the goal encoder is updated using EMA updates from the object encoder rather than direct gradient propagation.

As reported in \cref{tab:metrics_alignment_ablation}, our continuous contrastive objective achieves the best downstream control performance, with the highest number of CS and the lowest mean TTS. We attribute this to the continuous nature of the alignment objective: shaping embedding similarity according to angular distance provides a graded signal of task progress that is useful for closed-loop reorientation. The separate and EMA variants perform worse overall, indicating that when the object and goal encoders are decoupled, the policy must learn to compare poorly-matched representations, leading to lower task success. 
Shared encoders improve reconstruction accuracy, especially for pose-related quantities, likely because the shared visual backbone receives supervision from both object and goal image pathways. However, they do not achieve the best policy performance, highlighting that the best representations for reconstruction may not necessarily be the best for control. InfoNCE achieves the highest F1 score for success prediction and the lowest prediction failure rate, demonstrating that binary contrastive supervision is effective for classifying goal completion. However, its lower CS and higher TTS suggest that discrete threshold-based alignment provides a weaker signal for goal-directed control than our angular-distance-based objective.

\vspace{-2pt}
\subsection{Scaling Object Diversity}
\vspace{-2pt}

\credit{
We evaluate our approach on an additional dataset of 25 diverse objects by distilling a privileged teacher into an image-goal-conditioned student using the same architectures and training hyperparameters as for the 5-object dataset. As shown in~\cref{tab:metrics_scale}, the teacher achieves comparable performance on both datasets. We compare two student configurations: the original setup and a scaled setup with twice as many parallel environments (1280 vs. 640). The original setup achieves very limited task success on the larger dataset. As the number of objects increases, each object is sampled in fewer environments per iteration, potentially increasing gradient noise. Scaling up the training batch size (Image-Goal Conditioned 2x) substantially recovers student performance in both task success and reconstruction accuracy.
}

\vspace{-2pt}
\subsection{Real-World Deployment}
\vspace{-2pt}

\begin{figure*}[t]
  \centering

  \begin{minipage}[c]{0.546\textwidth}
    \centering

    \renewcommand{\arraystretch}{1.44} %
    \setlength{\tabcolsep}{5pt}        %

    \resizebox{\linewidth}{!}{%
      \begin{tabular}{
        >{\raggedright\arraybackslash}p{2.150cm}
        >{\centering\arraybackslash}p{1.7cm}
        >{\centering\arraybackslash}p{1.6cm}
        >{\centering\arraybackslash}p{1.6cm}
        >{\centering\arraybackslash}p{1.6cm}
      }
        \toprule
        \textbf{Object}
        & \textbf{Successes (Out of 20)}
        & \textbf{Object Drop}
        & \textbf{Timeout}
        & \textbf{Prediction Failure} \\
        \midrule
        Cube           &  18 & 0 & 0 & 2 \\
        3D Printed Toy &  16 & 1 & 2 & 1 \\
        Tablet Bottle  &  13 & 2 & 1 & 4 \\
        Rubber Duck    &  17 & 0 & 0 & 3 \\
        Globe          &  16 & 1 & 1 & 2 \\
        \midrule
        Mean
        & \(16.0 _{\pm 1.67}\)
        & \(0.8 _{\pm 0.75}\)
        & \(0.8 _{\pm 0.75}\)
        & \(2.4 _{\pm 1.02}\) \\
        \bottomrule
      \end{tabular}%
    }
    \vspace{-4pt}
  \captionof{table}{
    Real-world deployment results across five objects.
    Our image-goal-conditioned policy achieves an average success rate of \(80\%\) over 100 real-world trials.
    Prediction failures constitute the primary failure mode, suggesting that estimating goal completion remains the main limitation during deployment.
}
    \label{tab:hardware_deployment_table}
  \end{minipage}
  \hfill
  \begin{minipage}[c]{0.416\textwidth}
    \centering
    \includegraphics[width=\linewidth]{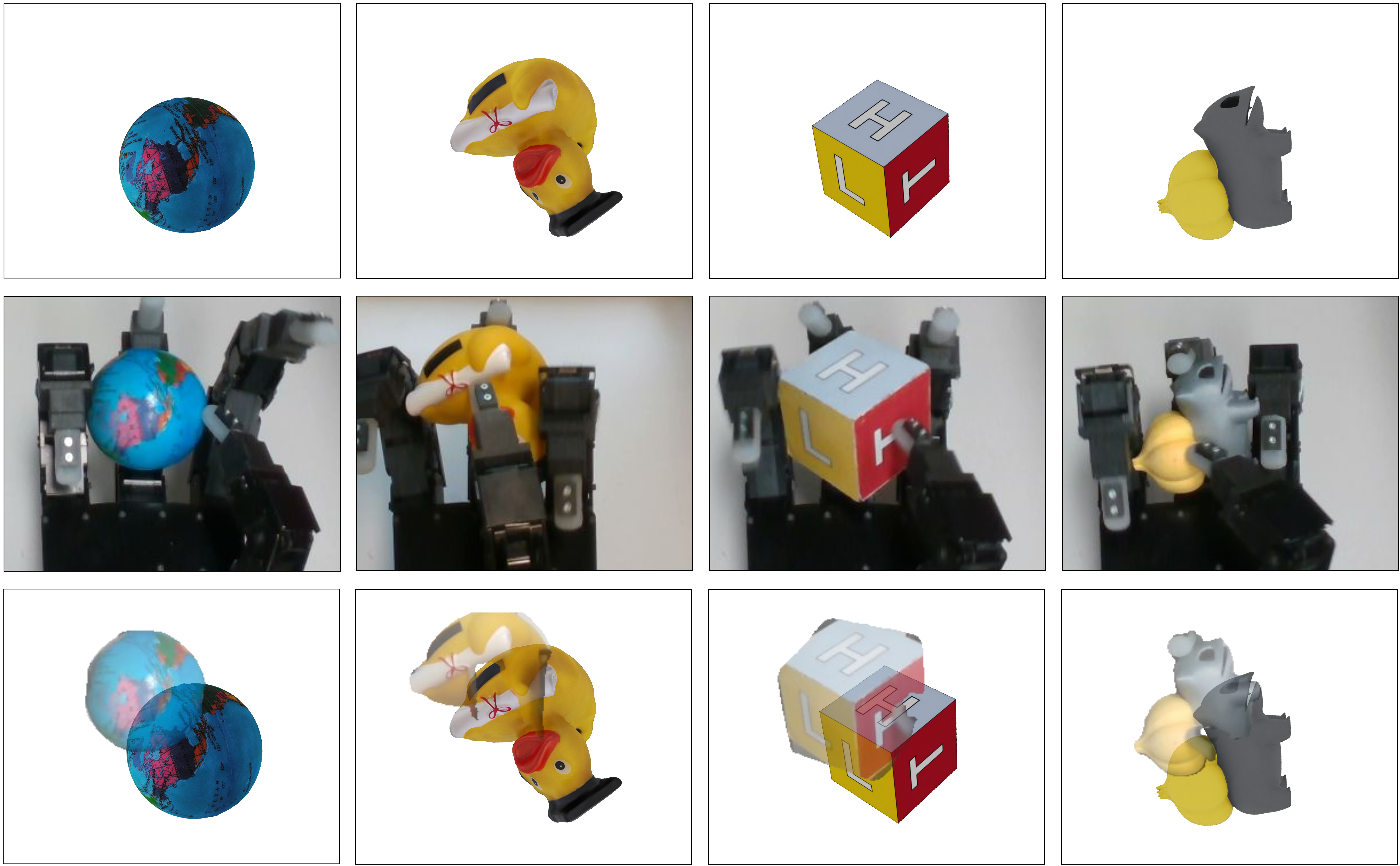}
    \vspace{-14pt}
    \captionof{figure}{Top: goal images \(I_g\). 
Center: our image-goal-conditioned policy successfully reaches the specified goals. 
Bottom: overlays of the current observation \(I_t\) and goal image \(I_g\), showing that successful configurations can still remain visually different.}
    \label{fig:hardware_deployment}
  \end{minipage}
\vspace{-18pt}
\end{figure*}

We evaluate a single image-goal-conditioned policy on real hardware across five objects with diverse geometries and physical characteristics. Each object is evaluated over 20 randomly sampled reorientation goals, with per-object successes and failure modes reported in~\cref{tab:hardware_deployment_table}. Our policy transfers zero-shot to real hardware, and achieves an average success rate of \(80.0\%\) across all evaluated objects. \credit{Under an identical protocol, the strongest pose-based baseline with class encodings succeeds in only 27 of the 100 trials.}
We further categorize failures into object drops, timeouts, and prediction failures. Prediction failures are the most common failure mode, while drops and timeouts occur infrequently. This indicates that the policy generally maintains stable manipulation, but that estimating goal completion from image-space observations remains a primary source of deployment error. The \emph{Tablet Bottle} is the most challenging object due to its symmetric geometry, low-friction surface, and visually ambiguous texture, which degrades both control stability and success prediction. Goal images and corresponding successful camera observations are shown in~\cref{fig:hardware_deployment}. \credit{We provide the procedure for real-world success-labeling in \cref{app:real-world-label}.}

\vspace{-2pt}
\subsection{Language-Guided Reorientation via Visual Goal Selection}
\vspace{-2pt}

\begin{figure*}[t]
  \centering
    \includegraphics[width=\textwidth]{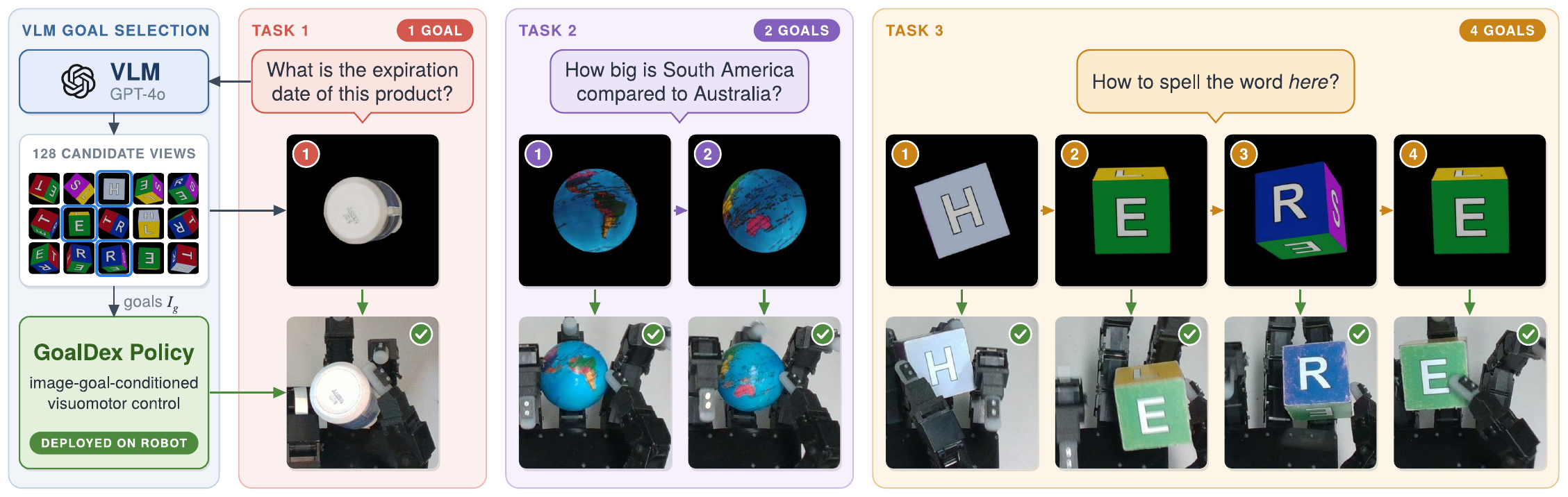}
    \vspace{-15pt}
    \captionof{figure}{
Language-guided reorientation via visual goal selection.
Given a task prompt, a vision-language model selects task-relevant goal images from candidate object viewpoints, which are then executed by the low-level controller.
Colored boxes show the task prompts, the middle row shows selected goal images, and the bottom row shows camera observations after successful reorientation.
}
\vspace{-18pt}
    \label{fig:vlm_integration}
\end{figure*}

\credit{We demonstrate that image-goal-conditioned policies can interface with vision-language models (VLMs) for task-oriented object reorientation. Since our joint-level policy operates directly on image-space goals, high-level reasoning models can specify manipulation targets by selecting appropriate goal images, providing a natural interface between internet-scale visual reasoning and dexterous control.} \credit{We evaluate language-guided reorientation on three tasks: locating a product’s expiration date, comparing the size of two continents on a globe, and spelling \emph{here} by sequentially exposing the letters \emph{H}, \emph{E}, \emph{R}, and \emph{E}. For each task, the VLM selects the relevant goal image or sequence of images, and the learned image-goal conditioned policy executes the corresponding reorientation. The selected goal images and the corresponding camera observations after successful reorientation are shown in~\cref{fig:vlm_integration}. These results highlight image-space goals as a flexible interface between high-level semantic reasoning and low-level joint control, requiring the VLM to select visual targets rather than predict explicit poses or robot commands.}

\vspace{-3pt}
\section{Limitations}
\label{sec:limitations}
\vspace{-3pt}

Our approach demonstrates image-goal-conditioned reorientation across multiple objects, but limitations remain. The current training is limited to 25 objects and does not yet generalize to entirely unseen object categories. Scaling to broader object distributions will likely require more diverse training data and representation learning methods that combine internet-scale visual priors with interaction-aware dynamics learned from robotic experience. Our goal-image conditioning also assumes access to object geometry for rendering target views. Future work could relax this assumption by using target images retrieved from similar objects, or by leveraging object reconstruction and generative 3D models to synthesize goal views. In addition, removing the dependence on segmented object masks through full-frame end-to-end training or jointly learned segmentation would simplify real-world deployment. Finally, our hardware experiments are performed in a palm-up hand configuration. Although the method is not fundamentally tied to this setup, extending it to more general manipulation scenarios will require coupling the dexterous policy with arm motion.

\vspace{-3pt}
\section{Conclusion}
\label{sec:conclusion}
\vspace{-3pt}

We present an end-to-end visuomotor framework for dexterous in-hand reorientation that removes reliance on canonical object coordinate frames or intermediate pose estimators. Our policy learns dynamic, interaction-aware representations directly from monocular RGB observations and proprioceptive feedback. We enable object-frame-agnostic dexterous reorientation via image-goal conditioning, allowing manipulation targets to be specified as visual goal observations. To align the current and goal visual representations for efficient learning, we introduced a continuous contrastive objective that structures the latent space along a rotational path toward the target configuration.

Extensive simulation experiments demonstrate that our approach outperforms pose-based pipelines, frozen pretrained visual encoders, and architectural ablations across diverse object geometries and scales. Finally, we demonstrate direct sim-to-real transfer of a single image-goal-conditioned policy across five geometrically distinct real-world objects. Overall, our results suggest that dexterous manipulation policies can move beyond explicit pose abstractions by learning task-relevant visuomotor representations directly from sensory observations. We believe that image-space goal conditioning provides a promising interface between low-level dexterous control and higher-level visual reasoning, opening a path toward more general, object-agnostic dexterous manipulation systems.

\clearpage

\section*{Acknowledgments}
This work was funded by ETH Zurich (research grant \mbox{no. 22-2 ETH-47}), Swiss National Science Foundation (NCCR Automation grant \mbox{no. 51NF40\_225155}), Swiss Federal Railways (SBB) via ETH Mobility Initiative, ETHAR, and by grants from NVIDIA and a donation from Meta. The authors also acknowledge the use of NVIDIA RTX 6000 Ada Generation GPUs, which facilitated this research. The authors would like to thank René Zurbrügg for insightful discussions on representation learning and Kaixian Qu for providing feedback on VLM integration.

\bibliography{references}  %

\clearpage

\appendix
\section{Additional Results}

\subsection{Privileged and Oracle-Pose Policies}
\label[appendix]{app:oracle-baselines}

\providecommand{\colspacing}{3pt}
\begin{table}[h]
\centering
\renewcommand{\arraystretch}{1.55}
\resizebox{0.9\textwidth}{!}{%
  \setlength{\tabcolsep}{\colspacing} %
  \footnotesize %
  \begin{tabular}{lcccccccc|cccc}
\hline
\textbf{\shortstack{\rule{0pt}{2.5ex}~ \\ Method \\ ~\rule[-1.5ex]{0pt}{0pt}}} &
\textbf{\shortstack{\rule{0pt}{2.5ex}~ \\ JP \\ ~\rule[-1.5ex]{0pt}{0pt}}} &
\textbf{\shortstack{\rule{0pt}{2.5ex}~ \\ OP \\ ~\rule[-1.5ex]{0pt}{0pt}}} &
\textbf{\shortstack{\rule{0pt}{2.5ex}~ \\ C \\ ~\rule[-1.5ex]{0pt}{0pt}}} &
\textbf{\shortstack{\rule{0pt}{2.5ex}~ \\ S \\ ~\rule[-1.5ex]{0pt}{0pt}}} &
\textbf{\shortstack{\rule{0pt}{2.5ex}~ \\ F \\ ~\rule[-1.5ex]{0pt}{0pt}}} &
\textbf{\shortstack{\rule{0pt}{2.5ex}~ \\ I \\ ~\rule[-1.5ex]{0pt}{0pt}}} &
\textbf{\shortstack{\rule{0pt}{2.5ex}~ \\ GP \\ ~\rule[-1.5ex]{0pt}{0pt}}} &
\textbf{\shortstack{\rule{0pt}{2.5ex}~ \\ GI \\ ~\rule[-1.5ex]{0pt}{0pt}}} &
\textbf{\shortstack{\rule{0pt}{2.5ex}Consecutive \\ Successes \\ $\uparrow$\rule[-1.5ex]{0pt}{0pt}}} &
\textbf{\shortstack{\rule{0pt}{2.5ex}Time to \\ Success \\ (s)~$\downarrow$\rule[-1.5ex]{0pt}{0pt}}} &
\textbf{\shortstack{\rule{0pt}{2.5ex}Success \\ Prediction \\ F1 Score~$\uparrow$\rule[-1.5ex]{0pt}{0pt}}} &
\textbf{\shortstack{\rule{0pt}{2.5ex}Prediction \\ Failure \\ Rate (\%)~$\downarrow$\rule[-1.5ex]{0pt}{0pt}}} \\
\hline
\multicolumn{7}{l}{\emph{Oracle Pose Policies}} \\
Oracle Pose & \gcheck & \gcheck & \rcross & \rcross & \rcross & \rcross & \gcheck & \rcross & \cellcolor[HTML]{EDCCD3} $22.49_{\pm 6.77}$ & \cellcolor[HTML]{EDCCD3} $5.38_{\pm 0.83}$ & \cellcolor[HTML]{EDCCD3} $0.782_{\pm 0.000}$ & \cellcolor[HTML]{FEEADD} $11.13_{\pm 0.00}$ \\
Oracle Pose + Class & \gcheck & \gcheck & \gcheck & \rcross & \rcross & \rcross & \gcheck & \rcross & \cellcolor[HTML]{FDE3DA} $43.77_{\pm 7.06}$ & \cellcolor[HTML]{FEF2E2} $4.43_{\pm 0.41}$ & \cellcolor[HTML]{FEF8E7} $\underline{0.869_{\pm 0.002}}$ & \cellcolor[HTML]{EDCCD3} $13.04_{\pm 1.19}$ \\
Oracle Pose + Scale & \gcheck & \gcheck & \rcross & \gcheck & \rcross & \rcross & \gcheck & \rcross & \cellcolor[HTML]{F9DBD6} $37.05_{\pm 0.00}$ & \cellcolor[HTML]{FEF2E2} $4.43_{\pm 0.00}$ & \cellcolor[HTML]{FEF8E7} $0.868_{\pm 0.000}$ & \cellcolor[HTML]{FDE5DB} $11.63_{\pm 0.00}$ \\
Oracle Pose + Class + Scale & \gcheck & \gcheck & \gcheck & \gcheck & \rcross & \rcross & \gcheck & \rcross & \cellcolor[HTML]{E5F3E0} $98.64_{\pm 0.00}$ & \cellcolor[HTML]{FEFDEF} $4.08_{\pm 0.00}$ & \cellcolor[HTML]{D2EBDC} $0.952_{\pm 0.000}$ & \cellcolor[HTML]{DBEFDE} $\underline{3.13_{\pm 0.00}}$ \\
\hline
\multicolumn{7}{l}{\emph{Privileged Policies}} \\
Privileged Student (Distilled) & \gcheck & \gcheck & \gcheck & \gcheck & \gcheck & \rcross & \gcheck & \rcross & \cellcolor[HTML]{CDE3D8} $\underline{120.38_{\pm 7.85}}$ & \cellcolor[HTML]{E1F2DF} $\underline{3.20_{\pm 0.44}}$ & \cellcolor[HTML]{CCE0D7} $\mathbf{1.000_{\pm 0.000}}$ & \cellcolor[HTML]{CCE0D7} $\mathbf{2.28_{\pm 0.35}}$ \\
Privileged Teacher (RL)       & \gcheck & \gcheck & \gcheck & \gcheck & \gcheck & \rcross & \gcheck & \rcross & \cellcolor[HTML]{CCE0D7} $\mathbf{122.96_{\pm 2.38}}$ & \cellcolor[HTML]{CCE0D7} $\mathbf{2.61_{\pm 0.62}}$ & - & - \\
\hline
\multicolumn{7}{l}{\emph{Image-Based Policy}} \\
Image-Goal Conditioned (Ours) & \gcheck & \rcross & \rcross & \rcross & \rcross & \gcheck & \rcross & \gcheck & \cellcolor[HTML]{FBFDED} $77.30_{\pm 3.25}$ & \cellcolor[HTML]{FEF6E5} $4.35_{\pm 0.24}$ & \cellcolor[HTML]{F3FAE5} $0.921_{\pm 0.000}$ & \cellcolor[HTML]{FEFCEE} $8.80_{\pm 0.15}$ \\
\hline
\end{tabular}%
}
\vspace{8pt}
\caption{Comparison with oracle-pose and privileged policies.
JP: joint proprioception; OP: oracle object pose; C: object class encoding;
S: object size; F: teacher's privileged observations; I: current RGB image;
GP: pose-space goal; GI: image-space goal.
}
\label{tab:oracle_pose_baselines}
\end{table}

\Cref{tab:oracle_pose_baselines} evaluates privileged student policies trained with joint states and oracle pose (OP), progressively augmented with class encodings (C), object scale (S), and the teacher's full observation set (F). The oracle-pose policy alone does not outperform our RGB policy, indicating that the performance gap in Table~1 cannot be attributed solely to errors from the learned pose estimator. This is consistent with pose specifying the object's spatial configuration but not its instance-level geometry. Adding class encodings and scale enables the policy to surpass our RGB policy in consecutive successes, highlighting the value of object-specific information. Our RGB policy implicitly recovers such information from visual observations and joint feedback, achieving approximately $63\%$ of the teacher's consecutive successes despite partial observability. Finally, the student trained with the full privileged observation set matches the teacher's performance. Importantly, the oracle-pose policy augmented with class encodings provides an optimistic reference for deployable pose-based approaches, since object scale and the teacher's full privileged state are not directly available from monocular observations.

\subsection{Per-Object Metrics}
\label[appendix]{app:object-metric}

\begin{figure*}[t]
  \centering
  \begin{subfigure}[b]{0.49\textwidth}
    \centering
    \includegraphics[width=\linewidth]{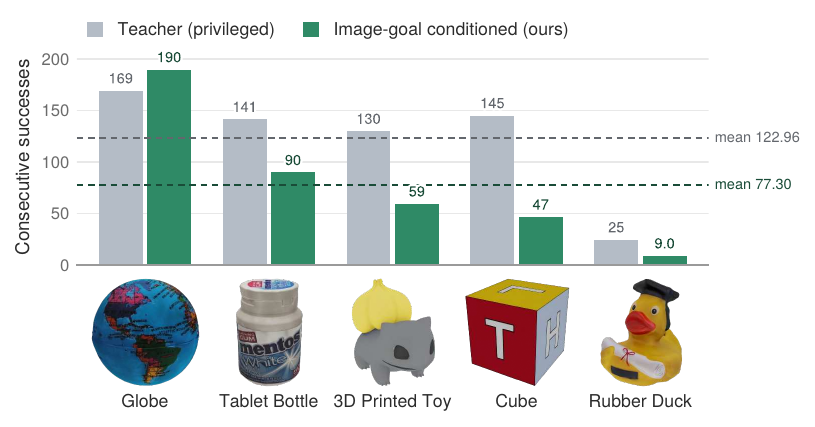}
    \caption{Consecutive successes, five-object set}
    \label{fig:per_object_5_cs}
  \end{subfigure}\hfill
  \begin{subfigure}[b]{0.49\textwidth}
    \centering
    \includegraphics[width=\linewidth]{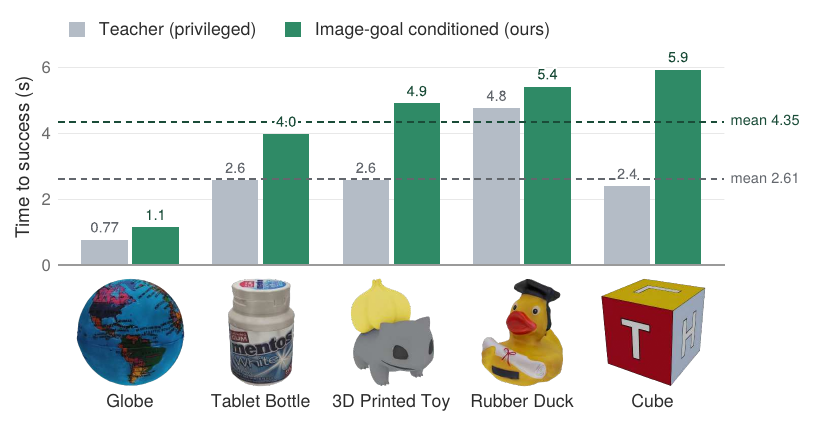}
    \caption{Time to success, five-object set}
    \label{fig:per_object_5_tts}
  \end{subfigure}

  \vspace{0.6em}
  \begin{subfigure}[b]{\textwidth}
    \centering
    \includegraphics[width=\linewidth]{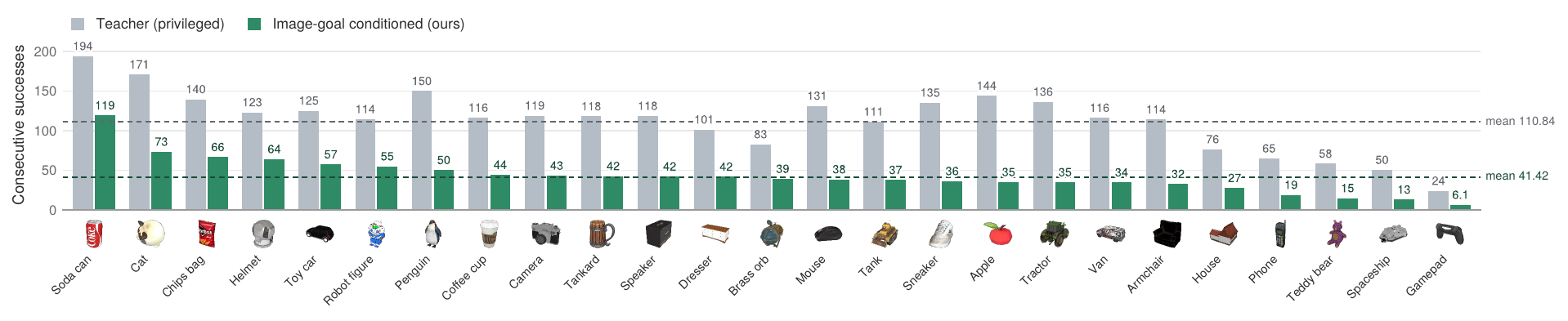}
    \caption{Consecutive successes, 25-object set}
    \label{fig:per_object_25_cs}
  \end{subfigure}

  \vspace{0.6em}
  \begin{subfigure}[b]{\textwidth}
    \centering
    \includegraphics[width=\linewidth]{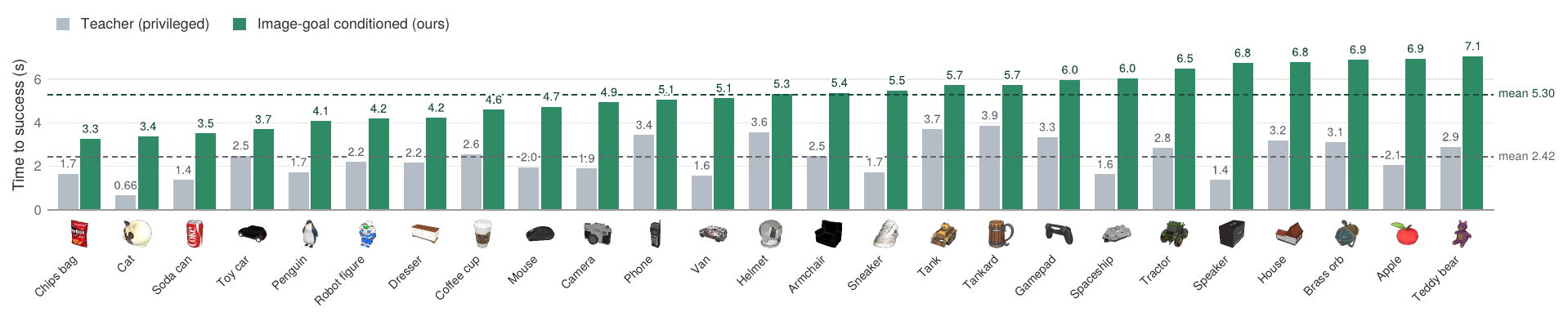}
    \caption{Time to success, 25-object set}
    \label{fig:per_object_25_tts}
  \end{subfigure}

  \caption{Per-object performance of the privileged teacher and our image-goal conditioned policy. Bars are
    per-object means; dashed lines mark each policy's mean over all objects. Objects are ordered by our policy's performance
    (consecutive successes: highest first; time to success: fastest first).
  }
  \label{fig:per_object}
\end{figure*}

We report per-object consecutive successes for the 5-object and 25-object datasets in \cref{fig:per_object}. The teacher maintains comparable performance across both sets, achieving 122.96 and 110.84 consecutive successes on average, respectively, with similar time per goal (2.42 s vs. 2.61 s). In contrast, student performance drops from 77.3 to 41.4 consecutive successes. This decline coincides with approximately 2.5$\times$ fewer training samples per object, suggesting that scaling the training setup with dataset size could help mitigate the performance gap.

Performance within each dataset varies with object geometry. Compact, rounded, and rotationally symmetric objects are generally easiest to reorient. On the globe, our policy matches the teacher and reaches each goal in approximately one second; in the 25-object dataset, the soda can, cat, and chips bag achieve the highest scores for both policies. Thin, flat, or elongated objects and those with protrusions are more challenging, including the gamepad, spaceship, teddy bear, phone, and rubber duck, consistent with observations in prior work~\cite{wang2024penspin}. Since the privileged teacher also struggles with these objects, their difficulty appears to stem primarily from manipulation rather than perception. By contrast, the student–teacher performance gap is more closely associated with perception than geometry: across the 25 objects, the student's performance relative to the teacher correlates with its success-prediction F1 score ($r = 0.62$).

\subsection{Standalone Pose Estimator Evaluation}
\label[appendix]{sec:pose_estimator_evaluation}

We evaluate the standalone accuracy of the pose estimator used to train pose-based baselines in ~\cref{tab:metrics_modality_ablation}. We report the position error, the geodesic orientation error, the error in the estimated angle to the goal $|\hat{d}_{\theta} - d_{\theta}|$, where $d_{\theta}$ is the angle between the current and target orientation and the success-prediction F1 score obtained by thresholding $\hat{d}_{\theta}$ at $0.4$~rad. The F1 score is computed the same as for the student policies.

\begin{table}[h]
    \centering
    \resizebox{0.7\columnwidth}{!}{
    \begin{tabular}{llcccc}
        \toprule
        \textbf{Model} & \textbf{Rollouts} & \shortstack{\textbf{Pos.} \\ \textbf{(mm) $\downarrow$}} & \shortstack{\textbf{Ori.} \\ \textbf{(rad) $\downarrow$}} & \shortstack{\textbf{Goal Angle} \\ \textbf{(rad) $\downarrow$}} & \shortstack{\textbf{Success} \\ \textbf{F1 $\uparrow$}} \\
        \midrule
        Pose estimator & {Teacher} & $6.76_{\pm 0.09}$ & $0.245_{\pm 0.015}$ & $0.153_{\pm 0.016}$ & $0.846_{\pm 0.027}$ \\
        Pose estimator & {Student} & $6.82_{\pm 0.18}$ & $0.250_{\pm 0.022}$ & $0.147_{\pm 0.022}$ & -- \\
        \midrule
        GoalDex recon.\ head & Teacher & $4.50_{\pm 0.00}$ & $0.151_{\pm 0.004}$ & $0.136_{\pm 0.003}$ & $0.921_{\pm 0.000}$ \\
        \bottomrule
    \end{tabular}
    }
    \vspace{6pt}
    \caption{Standalone evaluation of the pose estimator used to train pose-based baselines.}
    \label{tab:pose_estimator_accuracy}
\end{table}

\Cref{tab:pose_estimator_accuracy} summarizes the results. The pose estimator achieves a mean position error of 6.8 mm and a mean orientation error of 0.25 rad on both teacher and student rollouts, comparable to the rollout errors reported in ~\cite{bhardwaj2026viserdex}. The estimators are equally accurate on the rollouts of the teacher (training data) and its pose-based students. The low task performance of these student policies in \cref{tab:metrics_modality_ablation} is therefore not explained by a degradation of the pose estimate under their own state distribution.

\section{Implementation Details}
\label[appendix]{app:method-details}

\subsection{Teacher Policy Learning}

\paragraph{MDP Formulation:}
We adopt the goal-conditioned MDP formulation from~\citep{bhardwaj2026viserdex} for training the privileged teacher policy. Specifically, we use the same action space, where the policy predicts normalized joint position targets \(a_t\). that are scaled to the robot joint limits and smoothed using a low-pass exponential moving average (EMA),
\(\hat{a}_t = (1-\alpha)\hat{a}_{t-1} + \alpha a_t .\)
We also use the same reward terms, termination conditions, and performance-based curriculum introduced in~\citep{bhardwaj2026viserdex}. The reward combines a task term \(r_{\text{task}}\), which encourages progress toward the target object orientation, with regularization terms \(r_{\text{reg}}\), which penalize unstable or undesirable manipulation behavior.

We train a single privileged teacher across multiple objects and object instances, rather than separate object-specific teachers. To support this multi-object setting, the teacher receives privileged simulation observations grouped into proprioceptive, state, object-identity, and goal observations. We summarize these observation groups in~\cref{tab:teacher_obs_space}.

\paragraph{Training Details:}

The privileged teacher policy encodes the goal observations and state observations using separate MLP encoders. The goal observations are projected into a 64-dimensional latent representation, while the privileged-state observations (along with the object-class one-hot encodings) are projected into a 128-dimensional latent representation. These latent features are concatenated with the proprioceptive observations and passed to the actor and critic MLPs. The corresponding network dimensions and PPO~\cite{schwarke2025rslrl} hyperparameters are summarized in~\cref{tab:teacher_ppo_params}.

\subsection{Multi-Object Gaussian Rendering}
\label{sec:gaussian_rendering}
To generate photorealistic training observations for the visuomotor policies, we adopt the 3D Gaussian Splatting (3DGS) rendering pipeline and pre-rasterization augmentation strategy proposed in~\citep{bhardwaj2026viserdex}. We extend the rendering pipeline to support multi-object training. During rendering, the pipeline queries the simulation backend to determine the active object identity and selects the corresponding 3D Gaussian representation for rasterization.

To ensure consistency between randomized scales ($S_i \in \mathbb{R}^{4 \times 4}$) applied to objects within the simulation backend and the rendered observations, we dynamically scale the rendered object geometry. Instead of modifying the underlying spatial definitions of the static 3D Gaussian scenes, we apply the transformation directly to the camera view matrices. Given a canonical homogeneous point position $x$, the world-to-camera view matrix $V_i$, and an independent spatial scale transformation $S_i$, the composite coordinate projection evaluates as: $x_{cam} = V_i (S_i x) = (V_i S_i)x$. We exploit the associativity of these affine projections to evaluate a modified virtual viewpoint matrix $\tilde{V}_i = V_i S_i$. Passing $\tilde{V}_i$ directly into the rasterization loop ensures exact visual alignment with physical collision meshes while bypassing the need to recompute or duplicate high-dimensional Gaussians online.

\begin{table}[t]
    \centering
    \resizebox{0.8\columnwidth}{!}{
    \begin{tabular}{>{\raggedright\arraybackslash}p{5.5cm} c p{8.0cm}@{}}
        \toprule
        \textbf{Observation Term} & \textbf{Dim.} & \textbf{Description} \\
        \midrule

        \multicolumn{3}{l}{\textbf{Proprioceptive Observations} ($\boldsymbol{\mathcal{O}_{\text{prop}}}$)} \\
        \cmidrule{1-3}
        Joint Positions        & 16 & Measured robot joint angles. \\
        Action History         & 64 & Joint position commands from the last four time steps. \\
        Remaining Time         & 1  & Time remaining in the episode before timeout (seconds). \\
        \multicolumn{1}{l}{\textbf{Group Total}} & \textbf{81} & \\
        \midrule

        \multicolumn{3}{l}{\textbf{Goal Observations} ($\boldsymbol{\mathcal{O}_{\text{goal}}}$)} \\
        \cmidrule{1-3}
        Goal Position          & 3  & Reference position for object stability (constant). \\
        Goal Orientation       & 4  & Target object orientation (quaternion). \\
        Goal Quaternion Diff.  & 4  & Rotation representing the difference from object to goal. \\
        \multicolumn{1}{l}{\textbf{Group Total}} & \textbf{11} & \\
        \midrule

        \multicolumn{3}{l}{\textbf{System State Observations} ($\boldsymbol{\mathcal{O}_{\text{state}}}$)} \\
        \cmidrule{1-3}
        \multicolumn{2}{l}{\textbf{Reconstructed}} \\
        Object Position        & 3  & Ground-truth object position. \\
        Object Orientation     & 4  & Ground-truth object orientation (quaternion). \\
        Object Velocities      & 6  & Linear and angular velocities of the object. \\
        Object Scale           & 3  & Randomized object scale. \\
        Object Mass            & 1  & Randomized object mass. \\
        Object Forces          & 3  & Net contact force on the object. \\
        Joint Velocities       & 16 & Angular velocities of robot joints. \\
        Robot Scale            & 3  & Randomized robot scale. \\
        Joint Torques          & 16 & Actuator-applied joint torques. \\
        Actuator Gains         & 32 & Randomized joint stiffness and damping. \\
        \multicolumn{2}{l}{\textbf{Not Reconstructed}} \\
        Fingertip Forces       & 12 & Net contact forces at the four fingertips. \\
        Scene Gravity          & 3  & Gravity vector. \\
        External Forces        & 6  & External forces and torques applied on the object. \\
        EMA Parameter          & 1  & Randomized EMA parameter ($\alpha$). \\
        Action Delay           & 1  & Randomized simulated action delay. \\
        \multicolumn{1}{l}{\textbf{Group Total}} & \textbf{110} & \\
        \midrule

        \multicolumn{3}{l}{\textbf{Object Class Observations} ($\boldsymbol{\mathcal{O}_{\text{class}}}$)} \\
        \cmidrule{1-3}
        One-Hot Encoding       & 5  & One-hot encoding vector representing the object class. \\
        \multicolumn{1}{l}{\textbf{Group Total}} & \textbf{5} & \\
        \midrule
        
        \textbf{Total Observation Dim.} & \textbf{207} & \\
        \bottomrule
    \end{tabular}
    }
    \vspace{0.1cm}
    \caption{Observation Space of the Privileged Teacher Policy}
    \label{tab:teacher_obs_space}
\end{table}

\begin{table}[t]
    \centering
    \resizebox{0.5\columnwidth}{!}{
    \begin{tabular}{>{\raggedright\arraybackslash}p{5cm} c}
        \toprule
        \textbf{Parameter} & \textbf{Value} \\
        \midrule

        \multicolumn{2}{l}{\textbf{Network Architecture}} \\
        \cmidrule{1-2}
        Actor MLP Dims         & [1024, 1024, 1024, 1024] \\
        Critic MLP Dims        & [1024, 1024, 1024, 1024] \\
        Goal Observation Encoder Dims      & [256, 256]    \\
        Goal Observation Latent Dim        &   64        \\
        State Observation MLP Dims    & [256, 256]       \\
        State Observation Latent Dim  & 128              \\
        Activation Function    & ELU                     \\
        \midrule

        \multicolumn{2}{l}{\textbf{PPO Hyperparameters}} \\
        \cmidrule{1-2}
        Environments           & 24576                   \\
        Steps per Env          & 24                      \\
        Learning Epochs        & 5                       \\
        Mini-Batches           & 12                      \\
        Optimizer              & Adam                    \\
        Learning Rate          & 0.001                   \\
        LR Schedule            & Adaptive                \\
        Max Gradient Norm      & 1.0                     \\
        Discount Factor ($\gamma$) & 0.998               \\
        GAE Parameter ($\lambda$)  & 0.95                \\
        Value Loss Coefficient & 0.5                     \\
        Value Loss Clip Parameter         & 0.2                \\
        Desired KL Div.        & 0.01                    \\
        Initial Noise Std.     & 1.0                     \\
        Entropy Coefficient    & 0.002                   \\
        \bottomrule
    \end{tabular}
    }
    \vspace{0.14cm}
    \caption{Training Parameters for the Privileged Teacher Policy}
    \label{tab:teacher_ppo_params}
\end{table}

\subsection{End-to-End Visuomotor Student Learning}

\paragraph{Student Observations.}
The student policy receives only proprioceptive and image observations, without access to privileged object state. The image observations consist of object-masked RGB images of the current object configuration and the corresponding goal image. The complete student observation space is summarized in~\cref{tab:student_obs_space}.

\begin{table}[t]
    \centering
    \resizebox{0.8\columnwidth}{!}{
    \begin{tabular}{>{\raggedright\arraybackslash}p{5.5cm} c p{8.0cm}@{}}
        \toprule
        \textbf{Observation Term} & \textbf{Dim.} & \textbf{Description} \\
        \midrule

        \multicolumn{3}{l}{\textbf{\emph{Proprioceptive Observations}} ($\boldsymbol{\mathcal{O}_{\text{prop}}}$)} \\
        Joint Positions        & 16 & Measured robot joint angles. \\
        Action History         & 64 & Joint position commands from the last four time steps. \\
        Remaining Time         & 1  & Time remaining in the episode before timeout (seconds). \\
        \multicolumn{1}{l}{\textbf{Group Total}} & \textbf{81} & \\
        \midrule

        \multicolumn{3}{l}{\textbf{\emph{Image Observations}} ($\boldsymbol{\mathcal{O}_{\text{image}}}$)} \\
        Current RGB Image      & $120{\times}120{\times}3$ & 3DGS rendered RGB image of the current state. \\
        Goal RGB Image         & $120{\times}120{\times}3$ & 3DGS rendered RGB image of the target state. \\
        \multicolumn{1}{l}{\textbf{Group Total}} & {$\mathbf{120{\times}120{\times}6}$} & \\
        \midrule
        
        \textbf{Total Observation Dim.} & \multicolumn{2}{l}{\textbf{81 + Visual Image Inputs}} \\
        \bottomrule
    \end{tabular}
    }
    \vspace{0.1cm}
    \caption{Observation Space of the Student Policy}
    \label{tab:student_obs_space}
\end{table}

\paragraph{Online DAgger.}
We train the student using an online variant of DAgger~\citep{ross2011reduction}. To improve training stability and provide informative trajectories during early learning, actions are sampled from the privileged teacher policy with probability \(\beta\) (mixing ratio) and from the student policy with probability \(1-\beta\). The mixing ratio is initialized to a high value and exponentially decayed throughout training until the student fully controls the environment. We found this gradual transition from teacher-controlled to student-controlled rollouts to be important for stable and sample-efficient student policy learning.

\paragraph{Student Architecture.}
The perceptive student policy uses a recurrent, multi-headed architecture that maps sensory observations directly to robot joint targets. The architecture (as shown in~\cref{fig:pipeline}) processes two observation streams: proprioceptive state, consisting of robot joint information and action history, and exteroceptive visual input, consisting of RGB images of the current object configuration and the target goal configuration.

The visual stream is processed using a ResNet-34 backbone, which extracts a 512-dimensional feature vector from each image. The current object image \(O^{obj}_{image}\) and goal image \(O^{goal}_{image}\) are encoded as \(z^{obj}=f_{\theta}(O^{obj}_{image})\) and \(z^{goal}=f_{\phi}(O^{goal}_{image})\), respectively. These features are passed through a visual projection MLP and concatenated to produce the visual representation. In parallel, the proprioceptive observations \(O_{prop}\) are processed by a dedicated MLP encoder to produce a proprioceptive latent representation \(z^{prop}\). The visual and proprioceptive latents are then concatenated into a multimodal feature vector \(x_t = [z_t^{obj}, z_t^{goal}, z_t^{prop}]\), which is provided to the recurrent module.

The core of the recurrent module is a two-layer GRU~\citep{chung2014empirical} with 512-dimensional hidden state, which integrates multimodal observations over time to handle partial observability, occlusions, and delayed visual feedback. The recurrent latent state is passed through a latent encoder MLP. In parallel, the instantaneous multimodal feature vector $x_t$ is processed by a feature encoder MLP and added as a skip connection to the latent representation. This combined representation is used by the actor MLP to predict the continuous action distribution over normalized joint position targets. A separate reconstruction MLP uses the same belief representation to predict privileged simulation quantities, including object pose, velocities, and physical properties.

\paragraph{Reconstruction Targets.}
The reconstruction head predicts a subset of privileged object-state observations available in simulation. The reconstructed quantities are detailed in~\cref{tab:teacher_obs_space}. For simplicity, we predict a single concatenated reconstruction vector and optimize it using a mean-squared error loss, rather than defining separate losses for each individual quantity. To account for differences in scale across reconstructed variables, all targets are normalized using running estimates of the mean and standard deviation computed during teacher training. The reconstruction objective, therefore, does not require manually tuned loss weights for individual privileged quantities.

\paragraph{Training Setup.}
Student policies are trained in parallel across 640 simulation environments using two NVIDIA H100 GPUs (80~GB VRAM each), with 320 environments per GPU. For the 25-object dataset, we double the batch size and scale training to 1280 parallel environments across four GPUs. Training takes approximately 30 hours per policy. The architecture and training hyperparameters are detailed in~\cref{tab:student_train_params}.

\begin{table}[t]
    \centering
    \resizebox{0.5\columnwidth}{!}{
    \begin{tabular}{>{\raggedright\arraybackslash}p{5.4cm} c}
        \toprule
        \textbf{Parameter} & \textbf{Value} \\
        \midrule

        \multicolumn{2}{l}{\textbf{Network Architecture}} \\
        \cmidrule{1-2}
        Actor MLP Dims                  & [1024, 1024, 512, 512] \\
        Reconstruction MLP Dims         & [1024, 1024, 512, 512] \\
        Proprioceptive Encoder Dims     & [1024, 1024]           \\
        Proprioceptive Latent Dim       & 512                    \\
        Visual Encoder Backbone         & ResNet-34              \\
        Visual Encoder Latent Dim       & 512 (per image)        \\
        Visual Encoder MLP Dims         & [1024, 1024, 512, 512] \\
        Recurrent Layer                 & GRU                    \\
        Recurrent Network Dims          & 512 (2 layers)         \\
        Latent Encoder Dims             & [1024, 1024]           \\
        Feature Encoder Dims            & [1024, 1024]           \\
        Activation Function             & ELU                    \\
        \midrule

        \multicolumn{2}{l}{\textbf{Training Hyperparameters}} \\
        \cmidrule{1-2}
        Environments                    & 640                   \\
        Steps per Env                   & 45                     \\
        Learning Epochs                 & 4                      \\
        Mini-Batches                    & 32                     \\
        Optimizer                       & Adam                   \\
        Learning Rate                   & 0.0003                 \\
        Initial Noise Std.              & 0.02                   \\
        Contrastive Loss Coefficient    & 0.1                    \\
        Contrastive Similarity Scale ($\alpha$)         & 2.5    \\
        Reconstruction Loss Coefficient        & 0.2                    \\
        Action Mixing Ratio (Initial)   & 0.9                    \\
        Action Mixing Decay Rate        & 0.995                  \\
        \bottomrule
    \end{tabular}
    }
    \vspace{0.14cm}
    \caption{Architecture and Training Parameters for the Student Policy}
    \label{tab:student_train_params}
\end{table}

\subsection{Deployment}

We use a 16-DOF Allegro Hand equipped with a wrist-mounted Intel RealSense D435i camera for visual observations. All models are deployed on a single workstation with an Intel Core i9 CPU and an NVIDIA RTX 6000 Ada GPU. The policy operates at \(30~\si{\hertz}\) and outputs target joint positions, which are tracked by a low-level joint PD controller running at \(300~\si{\hertz}\). Since the visual encoder operates on segmented object observations, we generate object masks for each input image during deployment. To achieve this, we fine-tune SAM2~\citep{ravi2024sam2} to segment the robot and background, enabling prompt-free real-time object segmentation.

We use a ResNet-34~\citep{He2015DeepRL} as the visual encoder backbone, which is the most computationally expensive component of the policy architecture. At each policy timestep \(t\), the action head receives a unified feature vector \(x_t\) containing the object-image embedding \(z_t^{obj}\), the goal-image embedding~\(z_t^{goal}\), and the proprioceptive embedding \(z_t^{prop}\). To satisfy real-time deployment constraints, we execute the visual encoder asynchronously with respect to the rest of the policy pipeline. The policy therefore operates on \(x_t = [z^{obj}_{t'}, z^{goal}_{t}, z^{prop}_{t}],\)
where \(t'\) denotes the timestep of the most recently available object-image embedding \(z^{obj}\), implemented using a zero-order hold approximation. Since the goal image remains fixed during a rollout, its embedding \(z_t^{goal}\) can be computed once and reused throughout execution. To improve robustness to deployment-time latency, we model stochastic delays in the visual observations during training, mimicking the asynchronous execution behavior encountered on real hardware.

During deployment, task completion is inferred from the task completion head through its prediction of the relative orientation \(q_{rel}\). The policy declares successful completion when the estimated angular error \(\hat{d}_{\theta}(q_{rel})\) remains below the predefined success threshold for three consecutive timesteps.

\subsection{Real-World Success Labeling}
\label[appendix]{app:real-world-label}

For \cref{tab:hardware_deployment_table}, prediction failures are identified through post-hoc analysis of recorded trials. For each frame in which the policy predicts success, we estimate the object pose using FoundationPose~\cite{Wen2023FoundationPoseU6} with high-resolution RGB-D data, object masks, known meshes, and multiple refinement iterations. Each estimate is then manually verified against a rendered overlay of the object mesh. This offline localization procedure is distinct from the online tracking required for real-time control. Goal poses are known a priori, as goal images are rendered from the object mesh at sampled poses. A trial is labeled as a \emph{prediction failure} if the policy predicts success but the measured relative orientation error exceeds the success threshold. A \emph{timeout} is recorded if the policy fails to achieve a successful reorientation within 10 seconds.

\subsection{Visual Goal Selection}

Given an object and its mesh, we first sample a discrete set of 128 candidate goal images from different viewpoints. For each task, we query a vision-language model (VLM)~\citep{achiam2023gpt} with the language instruction and a candidate image set. The VLM is asked to select the image, or ordered sequence of images, that would best satisfy the task. We use the following query format:
\begin{quote}
\small
You will be provided with a sequence of \(N\) images. They correspond to indices \(0\) through \(N-1\) in the exact order they are presented.

\textbf{Query:} \texttt{\{task prompt\}}

Analyze the images and select the index, or sequence of indices, of the best image(s) that can be used to answer the query. If the query implies a sequence of images where order matters, provide the ordered list of indices. Respond with a JSON object containing two keys: \texttt{selected\_indices}, a list of integers, and \texttt{reasoning}, a one- to two-sentence explanation.
\end{quote}

The selected images are then provided as image-space goals to our dexterous policy, which executes the corresponding in-hand reorientation sequence on the robot.

\section{Pose Estimator Baseline: Training Details}

\label[appendix]{app:baselines}
\label[appendix]{sec:pose_estimator}

The pose-based baselines in~\cref{tab:metrics_modality_ablation} use an estimated object pose as input to the policy. The pose is predicted by a standalone image-based pose estimator that receives the same rendered observation as the image-based students.
To keep the model capacity comparable to the image-based students, it uses the same backbone architecture, ResNet-34~\citep{He2015DeepRL}, initialized with ImageNet weights and fine-tuned during training. The head regresses nine keypoints where each keypoint is parameterized by its normalized image coordinates $(u, v)$ and its depth $d$, similar to \cite{bhardwaj2026viserdex}. The architecture and training parameters are listed in~\cref{tab:pose_estimator_train_params}.

\begin{table}[h]
    \centering
    \resizebox{0.5\columnwidth}{!}{
    \begin{tabular}{>{\raggedright\arraybackslash}p{5.4cm} c}
        \toprule
        \textbf{Parameter} & \textbf{Value} \\
        \midrule

        \multicolumn{2}{l}{\textbf{Network Architecture}} \\
        \cmidrule{1-2}
        Input                           & Object-masked RGB, $120{\times}120$ \\
        Visual Encoder Backbone         & ResNet-34 \\
        Backbone Input Resolution       & $224{\times}224$       \\
        Head MLP Dims                   & [1024, 512]            \\
        Activation Function             & ReLU                   \\
        \midrule

        \multicolumn{2}{l}{\textbf{Training Hyperparameters}} \\
        \cmidrule{1-2}
        Environment Steps               & 135{,}000              \\
        Mini-Batches per Update         & 32                     \\
        Gradient Steps                  & ${\sim}$226k           \\
        Loss                            & Huber ($\delta = 0.05$) \\
        Optimizer                       & AdamW                  \\
        Learning Rate                   & 0.0001                 \\
        Weight Decay                    & 0.01                   \\
        Image Augmentations             & ISO noise, motion blur \\
        \bottomrule
    \end{tabular}
    }
    \vspace{0.14cm}
    \caption{Architecture and Training Parameters for the Pose Estimator}
    \label{tab:pose_estimator_train_params}
\end{table}

\end{document}